\documentclass[sigconf,screen]{acmart}

\AtBeginDocument{%
  \providecommand\BibTeX{{%
    \normalfont B\kern-0.5em{\scshape i\kern-0.25em b}\kern-0.8em\TeX}}}

\acmConference[]{}{}{}
\acmBooktitle{}
\acmPrice{15.00}
\acmISBN{978-1-4503-XXXX-X/18/06}

\usepackage{amsmath,amsfonts,bm}

\def\Figref#1{Figure~\ref{#1}}

\def\eqref#1{equation~\ref{#1}}
\def\Eqref#1{Equation~\ref{#1}}

\def\1{\bm{1}}

\DeclareMathAlphabet{\mathsfit}{\encodingdefault}{\sfdefault}{m}{sl}
\SetMathAlphabet{\mathsfit}{bold}{\encodingdefault}{\sfdefault}{bx}{n}

\usepackage[utf8]{inputenc} 
\usepackage[T1]{fontenc}    
\usepackage{hyperref}       
\usepackage{url}            
\usepackage{booktabs}       
\usepackage{amsfonts}       
\usepackage{nicefrac}       
\usepackage{microtype}      
\usepackage{xcolor}         

\usepackage{caption}
\usepackage{subcaption}
\usepackage{graphicx}

\usepackage{multirow}
\usepackage{booktabs}
\usepackage{subcaption}
\usepackage{nicematrix,tikz,makecell}
\usepackage{wrapfig}
\usepackage{amsthm}

\usepackage{array,ragged2e,tabularx,multirow}

\usepackage{xspace}
\newcommand{\our}{CORE\xspace}
\usepackage{color, colortbl}
\definecolor{Gray}{gray}{0.9}

\newcommand{\kevin}[1]{\textcolor{red}{KD:~#1}}
\newcommand{\zhichun}[1]{\textcolor{red}{ZC:~#1}}

\renewcommand{\kevin}[1]{}
\renewcommand{\zhichun}[1]{}

\newtheorem{theorem}{Theorem}

\begin{document}

\title[\our: Data Augmentation for Link Prediction via Information Bottleneck]{\our: Data Augmentation for Link Prediction\\ via Information Bottleneck}

\author{Kaiwen Dong}
\affiliation{
\institution{University of Notre Dame}
\country{USA}}
\email{kdong2@nd.edu}

\author{Zhichun Guo}
\affiliation{
\institution{University of Notre Dame}
\country{USA}}
\email{zguo5@nd.edu}

\author{Nitesh V. Chawla}
\affiliation{
\institution{University of Notre Dame}
\country{USA}}
\email{nchawla@nd.edu}


\begin{abstract}
  Link prediction (LP) is a fundamental task in graph representation learning, with numerous applications in diverse domains. 
  However, the generalizability of LP models is often compromised due to 
  the presence of noisy or spurious information in graphs and the inherent incompleteness of graph data. 
  To address these challenges, we draw inspiration from the Information Bottleneck principle and propose a novel data augmentation method, COmplete and REduce (\our) to learn compact and predictive augmentations for LP models.
  In particular,
  \our aims to recover missing edges in graphs while simultaneously removing noise from the graph structures, 
  thereby enhancing the model's robustness and performance. 
  Extensive experiments on multiple benchmark datasets demonstrate the applicability and superiority of~\our over state-of-the-art methods,
  showcasing its potential as a leading approach for robust LP in graph representation learning.
\end{abstract}

\ccsdesc[500]{Information systems~Data mining}
\ccsdesc[500]{Computing methodologies~Regularization}

\keywords{link prediction, data augmentation, information bottleneck, graph neural networks}

\received{20 February 2007}
\received[revised]{12 March 2009}
\received[accepted]{5 June 2009}

\maketitle

\section{Introduction}
Graph-structured data is ubiquitous in various domains, including social networks~\cite{liben-nowell_link_2003}, recommendation systems~\cite{koren_matrix_2009}, and protein-protein interactions~\cite{szklarczyk_string_2019}. 
Link prediction (LP), the task of predicting missing or future edges in a graph, is a fundamental problem in graphs.
Over the years, a plethora of link prediction algorithms have been proposed, ranging from heuristics-based link predictors~\cite{liben-nowell_link_2003,adamic_friends_2003,zhou_predicting_2009,katz_new_1953} to more sophisticated graph neural network (GNN) based methods~\cite{kipf_variational_2016,zhang_link_2018,pan_neural_2022}.

One major challenge in LP is the quality, reliability, and veracity of graph data. In many real-world scenarios, data collection can be difficult due to factors such as incomplete information, errors in data labeling, and noise introduced by measurement devices or human mistakes~\cite{chen_iterative_2020,zugner_adversarial_2018}.
As a consequence, the graph constructed based on the collected data may contain missing or erroneous edges. This can subsequently impact the performance of LP models.
Moreover, the excessive dependence on noisy graphs can impede the model's capability in distinguishing the real and spurious edges, 
which harms the generalizability of models.
Therefore, the question of preserving the robust learning capacity and generalizability of LP models on noisy graphs remains unresolved.

To mitigate the degradation of model performance on noisy data with inferior data quality,
data augmentation (DA) has emerged as a powerful technique by artificially expanding the training dataset with transformed versions of the original data instances, primarily in the field of computer vision~\cite{li_data_2023,shorten_survey_2019}. 
However, in the context of LP, few works have been proposed to overcome the limitation of models on noisy graphs~\cite{zhao_graph_2023}.
For example, 
CFLP~\cite{zhao_learning_2022} employs causal inference 
by complementing counterfactual links into the observed graph. 
Edge Proposal~\cite{singh_edge_2021} seeks to inject highly potential edges into the graph as a signal-boosting preprocessing step. 
Nevertheless, these works fail to consider the inherent noise or that brought by the augmentation process, holding an implicit
assumption that the observed graph truly reflects the underlying relationships


In this paper, we investigate how to augment the graph data
for link prediction to accomplish two primary goals: 
\textbf{\textit{eliminating noise inherent in the data and recovering missing information in graphs.}}
To augment with robust, diverse, and noise-free data, 
we employ the Information Bottleneck (IB) principle~\cite{tishby_deep_2015,tishby_information_2000}. 
IB offers a framework for constraining the flow of information from input to output,
enabling the acquisition of a maximally compressed representation while retaining its predictive relevance to the task at hand~\cite{alemi_deep_2023}.
Learning such an effective representation is particularly appealing because it gives us a flexible DA pipeline where we can seamlessly integrate other DA techniques without concern for the introduction of extraneous noise they might bring.

\paragraph{Present work.} We introduce COmplete and REduce (\our), a novel data augmentation framework tailored for the link prediction task.~\our comprises of two distinct stages: the Complete stage and the Reduce stage. The Complete stage addresses the incompleteness of the graph by incorporating highly probable edges, resulting in a more comprehensive graph representation. One can plug in any link predictors that may be advantageous in recovering the graph's structural information, despite the possibility of introducing noisy edges.
The Reduce stage, which is the crux of the proposed method, operates on the augmented graph generated by the Complete stage. It aims to shrink the edge set while preserving those critical to the link prediction task. In doing so, the Reduce stage effectively mitigates any misleading information either inherently or introduced during the Complete stage. By adhering to the IB principle, the Reduce stage yields a minimal yet sufficient graph structure that promotes more generalizable and robust link prediction performance.

\begin{figure*}[t]
\begin{center}
\centerline{\includegraphics[width=\textwidth]{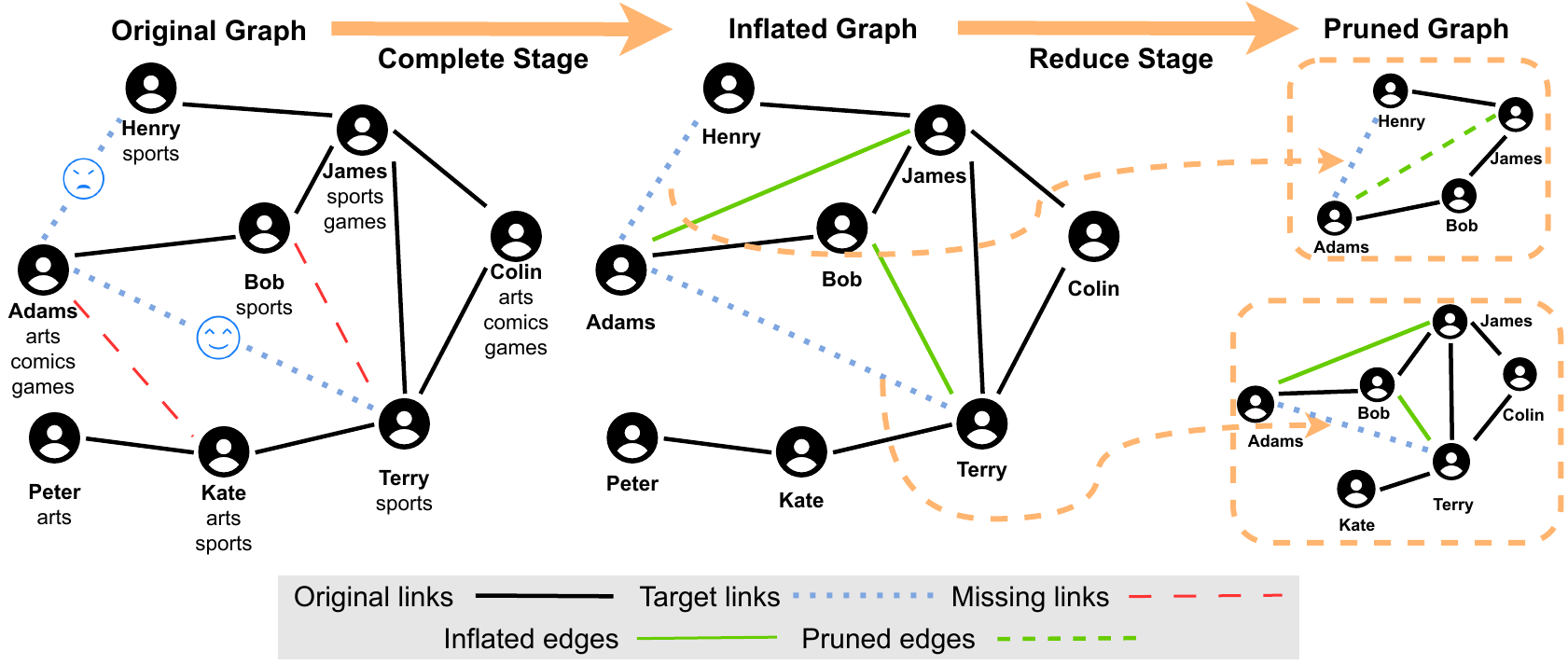}}
\caption{Overview of our~\our framework. It consists of two stages: 
(1) the Complete stage, which aims to recover missing edges by incorporating highly probable edges into the original graph, 
and (2) the Reduce stage, which is the core component of our method, 
designed to prune noisy edges from the graph 
in order to prevent overfitting on the intrinsic noise and that introduced by the Complete stage. 
Recognizing that predicting different links may require distinct augmentations, 
we extract the surrounding subgraph of each link and apply independent augmentations accordingly. 
In the social network example illustrated, 
assuming that Adams and Terry will become friends while Adams and Henry will not, 
tailored augmentations can facilitate more accurate link prediction by the model.}
\label{fig:overview}
\end{center}
\vspace{-2pt}
\end{figure*}

However, unlike DA in images where transformations can be applied independently, modifications to a single node or edge in a graph inevitably impact the surrounding neighborhood. This arises from the interdependence of data instances within a graph. 
Under these conditions, applying a universal DA to different instances in a graph may be suboptimal, as a specific augmentation could benefit one link prediction while negatively affecting another.
For example, in~\Figref{fig:overview}, the inference of different links may favor adding different edges into their neighborhood.
To address this dependency issue within a graph, we recast the link prediction task as a subgraph link prediction~\cite{zhang_link_2018,dong_fakeedge_2022}. 
In this context, we can apply different DAs to neighboring links without concerns about potential conflicts between their preferred augmentations. This approach allows for more targeted and effective augmentation, ultimately enhancing the performance of our~\our framework in link prediction tasks.



\section{Preliminary}
In this section, we introduce the notations and concepts utilized throughout the paper.

\paragraph{Graph and link prediction.}
Let $G = (V, E, \mathbf{X})$ represent an undirected graph, where $V$ is the set of nodes with size $n$, indexed as $\{i\}_{i=1}^{n}$.
$\mathcal{N}_v$ is the neighborhood of the node $v$.
$E \subseteq V \times V$ denotes the observed set of edges, and $\mathbf{X}_i \in \mathcal{X}$ represents the feature of node $i$.
The unobserved set of edges, denoted by $E_c \subseteq V \times V \setminus E$, comprises either missing edges or those expected to form in the future within the original graph $G$.
Thus, for those links $(i,j) \in E \cup E_c$, we can assign label $Y=1$ and regard them as positive samples, 
while the rest $\{(i,j) \subseteq V \times V | (i,j) \notin E \cup E_c\}$ we assign label $Y=0$ as negatives.
Based on the given graph $G$, the goal of the link prediction task is to 
compute the nodes similarity scores to identify the unobserved set of edges $E_c$~\cite{lu_link_2011}.
Numerous heuristic models are proposed for link prediction task over time, including Common Neighbor (CN)~\cite{liben-nowell_link_2003},
Adamic-Adar index (AA)~\cite{adamic_friends_2003}, Resource Allocation (RA)~\cite{zhou_predicting_2009}, and Katz index~\cite{katz_new_1953}. 
While these traditional approaches effectively utilize the topological structure of the graph,
GNN-based models~\cite{kipf_variational_2016,zhang_link_2018} exhibit a superior ability to exploit both the structure and node attributes associated with the graph.
\paragraph{Subgraph link prediction.}
Even though some link predictors, such as the Katz index and PageRank~\cite{brin_anatomy_1998}, require the entire graph to calculate similarity scores for a target link, 
many others only rely on a local neighborhood surrounding the target link for computation. 
For instance, the Common Neighbor predictor necessitates only a 1-hop neighborhood of the target link, 
and generally, an $l$-layer GNN requires the $l$-hop neighborhood of the target link. 
Moreover,~\citet{zhang_link_2018} has demonstrated that local information can be sufficient for link prediction tasks. 
As a result, the link prediction task for a specific target link can be reformulated as a graph classification problem based on the local neighborhood of the target link, 
aiming to determine whether the link exists or not~\cite{dong_fakeedge_2022}.
Formally, given a subgraph $G_{(i,j)}^{l}$ induced by the nodes $l$-hop reachable from node pair $(i, j)$, 
a subgraph link prediction is a task of predicting the label $Y \in \{0, 1\}$ for the subgraph, where $Y = 1$ indicates that the target link exists, and vice versa.

\paragraph{Data augmentation.} 
Data augmentation is the process of expanding the input data by either slightly perturbing existing data instances or creating plausible variations of the original data. 
This technique has been proven effective in mitigating overfitting issues during training, 
particularly in the fields of computer vision~\cite{cubuk_autoaugment_2019} and natural language processing~\cite{feng_survey_2021}. 
In the realm of graph representation learning, several DA methods have been proposed~\cite{zhao_graph_2023} to address challenges such as oversmoothing~\cite{rong_dropedge_2020}, generalization~\cite{chen_iterative_2020}, and over-squashing~\cite{topping_understanding_2022}. However, most graph data augmentation techniques have primarily focused on node and graph classification tasks, with relatively limited exploration in the context of LP~\cite{zhao_learning_2022}.



\section{Proposed framework:~\our}

In this section, we present our proposed two-stage data augmentation framework for LP, referred to as~\our. 
We begin by introducing the Complete stage, which aims to recover missing edges in the original graph. 
Following this, we discuss the Reduce stage, the most critical component of our proposed method, designed to eliminate noisy and spurious edges in the graph. 
Finally, we outline a practical implementation that leverages the Graph Information Bottleneck (GIB)~\cite{wu_graph_2020} for pruning inflated edges.
The overview of the framework is shown in~\Figref{fig:overview}.

\subsection{Complete stage: inflating missing connections}
The data collection process is inherently susceptible to errors, which can result in incomplete or even erroneous structural information in the original graph. 
Furthermore, the nature of the link prediction task involves identifying missing or potentially newly forming edges, 
implicitly assuming that graph data is incomplete. Therefore, mitigating the incompleteness of graph structures can be advantageous.
In Theorem~\ref{thm:reduce}, we will also see how inflating missing edges can help identify the most crucial component that determines whether a link should exist.

\paragraph{Implementation.}
We begin with a simple and straightforward method introduced by~\citet{singh_edge_2021} to inflate the original graph with additional edges. 
Although more sophisticated graph completion methods~\cite{wang_neural_2023} can also be plugged into the Complete stage, 
we find that employing a straightforward, low-computational-cost algorithm is sufficient for augmenting the graph structures effectively.

Due to the sparsity of most real-world graphs, the number of potentially missing edges is proportional to the quadratic of the number of nodes $\mathcal{O}(n^2)$. 
Scoring all non-connected node pairs can be computationally prohibitive. 
To reduce the size of the candidate node pairs, we only consider those non-connected pairs that have at least one common neighbor for potential addition to the graph.
Subsequently, we can use any link prediction method to score these candidate node pairs.
For large-scale datasets, like OGB-Collab~\cite{hu_open_2021}, 
we can rely on the faster computation of non-parametric heuristic methods.
For graphs of moderate size, one may choose any heuristic methods or GNN-based methods like GCN~\cite{kipf_semi-supervised_2017} and SAGE~\cite{hamilton_inductive_2018}. Note that only scoring node pairs with common neighbors is a pragmatic choice. Most real-world graphs, governed by popular models such as the assortative SBM~\cite{holland_stochastic_1983} or the Watts–Strogatz model~\cite{watts_collective_1998}, exhibit higher connection likelihood for node pairs with shared neighbors. Therefore, this design balances computational efficiency with empirical effectiveness, a trade-off we believe to be minimal but crucial for practical purposes. An empirical investigation can be found in Appendix \ref{app:only_common}.

After scoring all the candidate node pairs, 
we sort them based on their similarity scores and select the top $k$ node pairs $E_{ext}$ to add to the original graph, where $k$ is a hyperparameter.
Thus, the graph $G$ becomes $G^{+} = (V, E \cup E_{ext}, \mathbf{X})$.
It is important to note that, although we add these predicted links to the graph, we mark them as \emph{inflated edges} to differentiate them from the original graph.
These inflated edges will not be used as training signals for later stages but will only serve as a complement to the graph's topological structure. 
This distinction is crucial, as we can tolerate the noisy edges in the input space but do not want to introduce any noise to the labels of LP in our DA process.

\subsection{Reduce stage: pruning noisy edges}

\begin{figure*}[t]
\begin{center}
\centerline{\includegraphics[width=\textwidth]{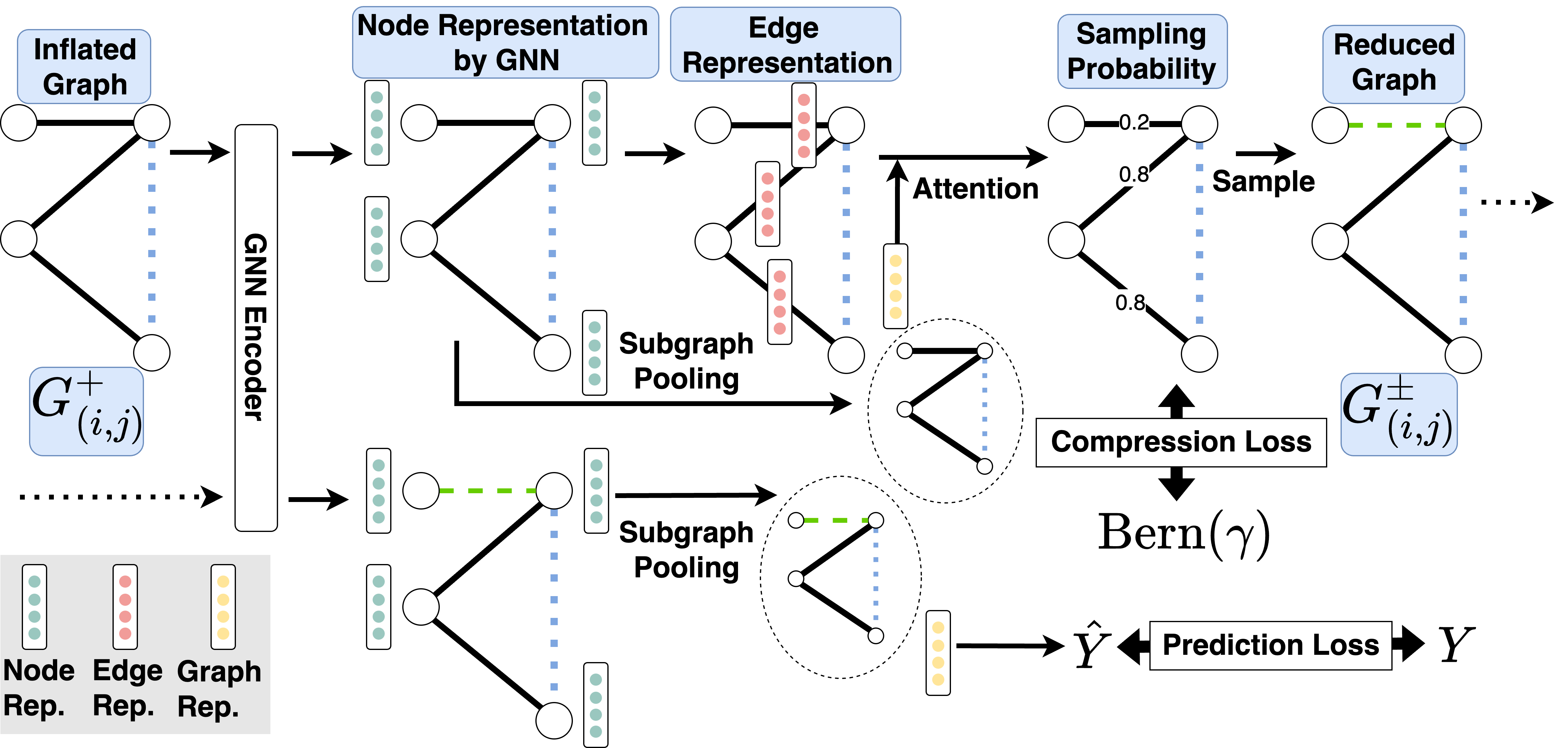}}
\caption{The Reduce stage commences with the inflated subgraph $G^{+}_{(i,j)}$ 
surrounding the target link $(i, j)$. 
We first apply a GNN to encode node representations, 
followed by edge representation derived from the node encodings. 
To compute sampling probability scores for each edge, 
we utilize an attention mechanism that combines the edge representation with the subgraph pooling. 
Since the subgraph pooling encapsulates information from the entire subgraph and is employed for target link prediction, 
the generated probability scores reflect not only the edge's inherent property
but also its relationship to the target link $(i, j)$. 
Subsequently, we sample each edge using a Bernoulli distribution based on its probability to obtain the pruned graph. 
Finally, the pruned graph $G^{\pm}_{(i,j)}$ is fed back into the model as augmented input for enhanced graph structures.
}
\label{fig:arch_reduce}
\end{center}
\vspace{-2pt}
\end{figure*}

The Reduce stage is the central aspect of our CORE data augmentation framework. 
Inspired by GSAT~\cite{miao_interpretable_2022}, we leverage GIB~\cite{wu_graph_2020} to
parameterize a reducer, which constrains the graph structure to 
a minimal yet sufficient graph component for link prediction. 
The resulting graph component is expected to achieve three goals: (1) remove task-irrelevant information from the data (the regularization in~\Eqref{eq:loss}); 
(2) prune the graph structures so that only the most predictive graph components remain for inference (the log-likelihood in~\Eqref{eq:loss}); 
and (3) provide diversified augmentation to the original data (the edge sampling step). 
We begin by introducing the necessity of decoupling the DA for each link. Then we discuss the GIB objective and its tractable variational bound. 
Finally, we present the implementation of our data augmentation in the Reduce stage.

\paragraph{Interdependence of graph data.}
In the link prediction task, the data instances we are interested in are the links in the graph. 
However, unlike images, links in a graph are correlated; the existence and properties of each link are dependent on one another. 
Consequently, when applying DA to a specific link, it will inevitably affect the environment of other links, especially those in close proximity. 
This can yield suboptimal results, as links will compete with each other to obtain the best augmentation for their own sake.
Furthermore, it becomes computationally infeasible to apply the IB principle when the i.i.d assumption does not hold~\cite{wu_graph_2020}.

To address these issues, we reformulate LP as a subgraph link prediction. 
Subgraph link prediction allows for decoupling the overlapping environments of each link, making it possible to have a different DA for each link. 
Specifically, for each node pair $(i, j)$, we extract its $l$-hop enclosing subgraph $G^{+;l}_{(i,j)}$ from the entire graph $G^+$. 
To simplify notation when there is no ambiguity, we may omit the number of hops and represent the subgraph as $G^{+}_{(i,j)}$. 
It is worth noting that prior works also adopt a similar strategy to handle the non i.i.d nature of graph data~\cite{wang_nodeaug_2020,zheng_robust_2020} when perturbing the data. In Section \ref{sec:edge_dis}, we empirically examine the optimal DAs tailored to different target links. Notably, the DA derived from a single edge may vary depending on the target link under consideration.

\paragraph{GIB}
In general, IB aims to learn a concise representation $Z$ from the input $X$ 
that is also expressive for the output $y$, measured by the mutual information between the latent representation and input/output~\cite{tishby_deep_2015,alemi_deep_2023}. 
Thus, the objective is:
\begin{align}\label{eq:IB}
\max_{Z} I(Z,Y) \text{ s.t. } I(X,Z) \leq I_c .
\end{align}
where $I(\cdot,\cdot)$ denotes the mutual information and $I_c$ is the information constraint.
In the context of LP, we can regard the enclosing subgraph $G_{(i,j)}^{+}$ as input $X$,
including both the node attributes and graph structure. 
$Y$ is the link's existence at $(i, j)$, and $Z$ is the latent representation.

While the original GIB~\cite{wu_graph_2020} constrains the information flow from both node attributes of a graph and graph structures, 
we propose to only constrain the structural information in our data augmentation for the link prediction task. 
Compared to node attributes in a graph, graph structures are overwhelmingly more critical for the link prediction task~\cite{pan_neural_2022,lu_link_2011}. 
Moreover, many graphs without node attributes still exhibit the need for link prediction. Thus, we define our objective as:
\begin{align}
\max_{G_{(i,j)}^{\pm} \in \mathbb{G}_{\text{sub}}(G_{(i,j)}^{+})} I(G_{(i,j)}^{\pm},Y) \text{ s.t. } I(G_{(i,j)}^{\pm},G_{(i,j)}^{+}) \leq I_c .
\end{align}
where $G_{(i,j)}^{\pm} \in \mathbb{G}_{\text{sub}}(G_{(i,j)}^{+})$ is a subgraph pruned from the inflated graph $G_{(i,j)}^{+}$. 
In other words, we aim to find the subgraph of the inflated graph that is simultaneously the most predictive and concise for the link prediction task.
We assume that this graph reduction process can prune the noisy edges introduced by the previous Complete stage while retaining the beneficial added information.
Our method shares a similar spirit with GSAT~\cite{miao_interpretable_2022} and IB-subgraph~\cite{yu_graph_2020}, as we explore finding a subgraph structure that is most essential for the task.

Next, by introducing a Lagrange multiplier $\beta$, we obtain the unconstrained version of the objective:
\begin{align}\label{eq:GIB}
\min_{G_{(i,j)}^{\pm} \in \mathbb{G}_{\text{sub}}(G_{(i,j)}^{+})} -I(G_{(i,j)}^{\pm},Y) + \beta I(G_{(i,j)}^{\pm},G_{(i,j)}^{+}).
\end{align}
where $\beta$ is the hyperparameter to balance the tradeoff between predictive power and compression.

The computation of the mutual information term $I(\cdot,\cdot)$ is, in general, computationally intractable. 
To address this issue, we follow the works of \citet{alemi_deep_2023,wu_graph_2020,miao_interpretable_2022} to derive a tractable variational upper bound for~\Eqref{eq:GIB}. 
The detailed derivation is provided in Appendix \ref{app:bound}.
To approximate the first term $I(G_{(i,j)}^{\pm}, Y)$, we derive a variational lower bound. The lower bound can be formulated as:
\begin{align} \label{eq:pred}
I( G_{(i,j)}^{\pm} ; Y)
\geq \mathbb{E}[ \log q_{\theta}(Y|G_{(i,j)}^{\pm}) ].
\end{align}
Essentially, $q_{\theta}$ is the predictor of our model, which can be a parameterized GNN.

For the second term $I(G_{(i,j)}^{\pm}, G_{(i,j)}^{+})$, we derive an upper bound by introducing a variational approximation $r(G_{(i,j)}^{\pm})$ for
the marginal distribution of $G_{(i,j)}^{\pm}$:
\begin{align} \label{eq:reg}
    I(G_{(i,j)}^{\pm},G_{(i,j)}^{+}) \leq \mathbb{E}[\text{KL}(p_{\phi}(G_{(i,j)}^{\pm} | G_{(i,j)}^{+})||r(G_{(i,j)}^{\pm})]
\end{align}
where $p_{\phi}$ is the reducer to prune noisy edges.
Then we can put everything together and get the empirical loss to minimize:
\begin{align} \label{eq:loss}
\mathcal{L} & \approx \frac{1}{|E|}\sum_{(i,j) \in E} [- \log q_{\theta}(Y|G_{(i,j)}^{\pm}) \\
& + \beta \text{KL}(p_{\phi}(G_{(i,j)}^{\pm} | G_{(i,j)}^{+})||r(G_{(i,j)}^{\pm}))].
\end{align}

Next, we discuss how to parameterize the predictor $q_{\theta}$ and the reducer $p_{\phi}$ in the Reduce stage, as well as the choice of marginal distribution $r(G_{(i,j)}^{\pm})$.


\subsection{Implementation of the Reduce stage.}
The overall architecture of the Reduce stage is shown in Figure \ref{fig:arch_reduce}. 
It is important to note that both the predictor $q_{\theta}$ and the reducer $p_{\phi}$ utilize 
a GNN encoder to encode the graph representation for either prediction or graph pruning purposes. 
These two components can share a common GNN encoder with the same parameters, as the entire training of the Reduce stage is end-to-end. 
This shared encoder allows for more efficient learning and reduces the number of parameters required in the model.

\paragraph{Subgraph encoding.}

The Reduce stage begins with encoding the inflated graph $G_{(i,j)}^{+}$ using a GNN. 
We can choose any Message Passing Neural Network (MPNN)~\cite{gilmer_neural_2017} as the instantiation of the GNN encoder. 
The MPNN can be described as follows:
\begin{align}\label{eq:mpnn}
\mathbf{m}_v^{(l)} = \text{AGG}\left(\{ \mathbf{h}_u^{(l)}, \mathbf{h}_v^{(l)}, \forall u \in \mathcal{N}_v \}\right),\\
\mathbf{h}_v^{(l+1)} = \text{UPDATE}\left( \{\mathbf{h}_v^{(l)}, \mathbf{m}_v^{(l)}\}\right).
\end{align}


where a neighborhood aggregation function $\text{AGG}(\cdot)$ and an updating function $\text{UPDATE}(\cdot)$ are adopted in the $t$-th layer of a $T$-layer GNN. Consequently, $\{\mathbf{h}_v^{(T)}| v \in G_{(i,j)}^{+}\}$ represents the node embeddings learned by the GNN encoder.

A typical link prediction method, such as GAE~\cite{kipf_variational_2016} or SEAL~\cite{zhang_link_2018}, 
can make a prediction by pooling the node representations, 
namely $\mathbf{h}_{G_{(i,j)}^{+}} = \text{POOL}\left({\mathbf{h}_v^{(T)}| v \in G_{(i,j)}^{+}}\right)$,
where $\mathbf{h}_{G_{(i,j)}^{+}}$ is the final representation for the node pair $(i,j)$.
In our data augmentation approach, however, we first need to prune the noisy edges in order to obtain more concise graph structures.
\paragraph{Reduce by edge sampling.}\label{para:edge_sampling}

After encoding the node representations, we proceed to prune the noisy edges in the inflated graph. 
We first represent each edge $(u,v)$ in the inflated graph $G_{(i,j)}^{+}$ by concatenating the node representations of the two end nodes and a trainable embedding 
indicating whether this edge comes from the original graph $G_{(i,j)}$ or the Complete Stage.
Specifically, we obtain $\mathbf{h}_{(u,v)} = \left[\mathbf{h}_u;\mathbf{h}_v;\tilde{\mathbf{h}}_{(u,v)}\right]$, 
where $\left[\cdot;\cdot\right]$ is the concatenation operation. 
Appending $\tilde{\mathbf{h}}_{(u,v)}$ to the edge representation enables the model to be aware of whether the edges are originally in the graph or introduced by link predictors at the Complete stage.

In this setting, the edge representation $\mathbf{h}_{(u,v)}$ solely contains information about 
its structural role in the inflated graph.
While structurally similar edges might influence distinct target node pairs differently,
this representation does not convey information about how the edge $(u,v)$ 
in the local subgraph $G_{(i,j)}^{+}$ affects the prediction of the target node pair $(i,j)$. 
To make the edge representation directly interact with the downstream link prediction task,
we further apply an attention mechanism~\cite{vaswani_attention_2017,velickovic_graph_2018} and 
attend it to the overall representation of the entire subgraph to 
define its importance for link prediction. We compute this as follows:
\begin{align}\label{eq:att}
a_{(u,v)} = { Q_\phi(\mathbf{h}_{G_{(i,j)}^{+}})^T K_\phi(\mathbf{h}_{(u,v)})} \big/ {\sqrt{F^{\prime \prime}}},
\end{align}
where $Q_\phi$ and $K_\phi$ are two MLPs and $F^{\prime \prime}$ is the output dimension of the MLPs.

Unlike GAT~\cite{velickovic_graph_2018}, which directly applies the attention scores as edge weights in each layer,
we use these scores $a_{(u,v)}$ to sample the edges to diversify the views of the graph. 
For each edge $(u,v)$ in $G_{(i,j)}^{+}$, we sample an edge mask from the Bernoulli distribution $\omega_{(u,v)}\sim \text{Bern}(\text{sigmoid}(a_{(u,v)}))$, 
which masks off unnecessary edges in the graph for LP.
To ensure the gradient can flow through the stochastic node here, 
we utilize the Gumbel-Softmax trick~\cite{jang_categorical_2023,maddison_concrete_2023}.
This procedure gives us a way to generate the reduced subgraph $G_{(i,j)}^{\pm}$ by
the variational distribution $p_{\phi}(G_{(i,j)}^{\pm} | G_{(i,j)}^{+})$.

To control the marginal distribution in~\Eqref{eq:reg}, 
we follow~\cite{miao_interpretable_2022,wu_graph_2020} 
and apply a non-informative prior $r(\tilde{G}_{(i,j)})$. In other words, $\tilde{G}_{(i,j)}$ is obtained by sampling edge connectivity $\tilde{\omega}_{(u,v)} \sim \text{Bern}(\gamma)$ for every node pair $(u,v)$ in $G_{(i,j)}^{+}$. $\gamma$ is a hyperparameter. Regardless the graph structure of $G_{(i,j)}^{\pm}$, we connect $(u,v)$ if $\tilde{\omega}_{(u,v)}=1$ and disconnect the rest. This is essentially an Erd\"os-R\'enyi random graph~\cite{erdos_random_1959}. The derivation of the KL loss term with respect to the marginal distribution is detailed in Appendix \ref{app:marginal}.
\paragraph{Prediction based on pruned subgraph.}
Once we obtain the edge mask $\omega$, we can encode the subgraph and make a link prediction. 
The edge mask can be regarded as the edge weight of the inflated graph. 
In this way, the edge weight plays the role of a message passing restrictor to prune the noisy edges in the inflated graph, which modifies the message passing part of~\Eqref{eq:mpnn} as follows:
\begin{align}\label{eq:mpnn_mask}
\mathbf{m}_v^{(l)} = \text{AGG}\left(\{ \omega_{(u,v)} * \mathbf{h}_u^{(l)}, \mathbf{h}_v^{(l)}, \forall u \in \mathcal{N}_v \}\right).
\end{align}
Using the representation learned from the reduced subgraph $G_{(i,j)}^{\pm}$, 
we can feed them into a pooling layer plus an MLP to estimate $Y$. 
This models the distribution $q_{\theta}(Y|G_{(i,j)}^{\pm})$.

\subsection{Theoretical analysis}
In this section, we provide a theoretical foundation for the integration of the Complete and Reduce stages in our data augmentation approach for link prediction tasks.
\begin{theorem}
\label{thm:reduce}
Assume that: \text{\normalfont (1)} The existence $Y$ of a link $(i,j)$ is solely determined 
by its local neighborhood
$G_{(i,j)}^{*}$ in a way such that $p(Y) = f(G_{(i,j)}^{*})$, 
where $f$ is a deterministic invertible function;
\text{\normalfont (2)} The inflated graph contains sufficient structures for prediction
$G_{(i,j)}^{*} \in \mathbb{G}_{\text{sub}}(G_{(i,j)}^{+})$.
Then $G_{(i,j)}^{\pm} = G_{(i,j)}^{*}$ minimizes the objective in~\Eqref{eq:GIB}.
\end{theorem}
The first assumption in Theorem~\ref{thm:reduce} is consistent with a widely accepted \emph{local-dependence}
assumption~\cite{wu_graph_2020,zhang_link_2018} for graph-structured data. 
The second assumption highlights the importance of incorporating enough structural information into the graph
in the Complete Stage prior to executing the reduce operation.
Even though we assume that the link existence $Y$ is causally determined by $G_{(i,j)}^{*}$,
there still can be some other spurious correlations between $Y$ and $G_{(i,j)}^{+}$.
These correlations can be brought by the environments~\cite{arjovsky_invariant_2019,krueger_out--distribution_2020,wu_handling_2022},
and the shift of such correlations in the testing phase can cause performance degradation for LP models.

Theorem~\ref{thm:reduce} implies that under mild assumptions,
optimizing the objective in~\Eqref{eq:GIB} can help us
uncover the most crucial component of the graph, which determines whether a link should exist.
As a result, our approach enables the elimination of noisy and spurious edges, thereby enhancing the generalizability of link prediction models.
The proof can be found in Appendix \ref{app:proof}.

\section{Experiments}
\begin{table*}[t]
\caption{Link prediction performance evaluated by Hits@50. 
The \textbf{best-performing} method is highlighted in bold, while the \underline{second-best} performance is underlined. OOM means out of memory.}
\label{table:main}
\begin{center}
  \resizebox{1\textwidth}{!}{
  \begin{NiceTabular}{ll|cccccccc}
    \toprule
    \textbf{Model Type} & \textbf{Models} & \textbf{C.ele} & \textbf{USAir} & \textbf{Yeast} & \textbf{Router} & \textbf{CS} & \textbf{Physics} & \textbf{Computers} & \textbf{Collab}\\
    
    \midrule
    \multirow{3}{*}{Heuristics}
    &\emph{CN} & 54.31{\tiny$\pm$0.00} & 82.59{\tiny$\pm$0.00} & 72.71{\tiny$\pm$0.00} & 9.11{\tiny$\pm$0.00} & 38.99{\tiny$\pm$0.00} & 63.44{\tiny$\pm$0.00} & 25.48{\tiny$\pm$0.00} & 61.37{\tiny$\pm$0.00} \\
    &\emph{AA} & 57.34{\tiny$\pm$0.00} & 87.53{\tiny$\pm$0.00} & 72.71{\tiny$\pm$0.00} & 9.11{\tiny$\pm$0.00} & 67.44{\tiny$\pm$0.00} & 74.38{\tiny$\pm$0.00} & 31.14{\tiny$\pm$0.00} & 64.17{\tiny$\pm$0.00} \\
    &\emph{RA} & 64.34{\tiny$\pm$0.00} & 87.53{\tiny$\pm$0.00} & 72.71{\tiny$\pm$0.00} & 9.11{\tiny$\pm$0.00} & 67.44{\tiny$\pm$0.00} & 74.68{\tiny$\pm$0.00} & 34.17{\tiny$\pm$0.00} & 63.81{\tiny$\pm$0.00} \\
    
    \midrule
    \multirow{3}{*}{\shortstack[l]{Network \\ Embedding}}
    &\emph{Node2Vec} & 50.82{\tiny$\pm$3.24} & 74.12{\tiny$\pm$2.12} & 82.11{\tiny$\pm$2.74} & 32.53{\tiny$\pm$4.23} & 63.32{\tiny$\pm$3.84} & 60.72{\tiny$\pm$1.85} & 28.48{\tiny$\pm$3.42} & 48.88{\tiny$\pm$0.54} \\
    &\emph{DeepWalk} & 48.62{\tiny$\pm$2.82} & 73.80{\tiny$\pm$1.98} & 81.24{\tiny$\pm$2.38} & 31.97{\tiny$\pm$3.92} & 64.18{\tiny$\pm$3.98} & 60.58{\tiny$\pm$2.24} & 27.49{\tiny$\pm$3.08} & 50.37{\tiny$\pm$0.34} \\
    &\emph{LINE} & 52.40{\tiny$\pm$2.02} & 74.82{\tiny$\pm$3.40} & 82.45{\tiny$\pm$2.75} & 34.39{\tiny$\pm$3.86} & 63.96{\tiny$\pm$2.83} & 61.90{\tiny$\pm$1.93} & 27.52{\tiny$\pm$2.98} & 53.91{\tiny$\pm$0.00} \\
    
    \midrule
    \multirow{5}{*}{GNNs}
    &\emph{GCN} & 57.32{\tiny$\pm$4.52} & 82.14{\tiny$\pm$1.99} & 80.33{\tiny$\pm$0.73} & 35.16{\tiny$\pm$1.60} & 60.69{\tiny$\pm$8.56} & 69.16{\tiny$\pm$4.61} & 32.70{\tiny$\pm$1.97} & 44.75{\tiny$\pm$1.07} \\
    &\emph{SAGE} & 42.14{\tiny$\pm$5.62} & 82.85{\tiny$\pm$4.01} & 78.34{\tiny$\pm$1.08} & 35.76{\tiny$\pm$2.97} & 31.44{\tiny$\pm$8.24} & 22.87{\tiny$\pm$22.53} & 14.53{\tiny$\pm$6.28} & 48.10{\tiny$\pm$0.81} \\
    &\emph{SEAL} & 67.32{\tiny$\pm$2.71} & 91.76{\tiny$\pm$1.17} & 82.50{\tiny$\pm$2.08} & 60.35{\tiny$\pm$5.72} & 65.23{\tiny$\pm$2.08} & 71.83{\tiny$\pm$1.44} & 35.80{\tiny$\pm$1.38} & 63.37{\tiny$\pm$0.69} \\
    &\emph{ELPH} & 66.06{\tiny$\pm$3.00} & 88.16{\tiny$\pm$1.21} & 78.92{\tiny$\pm$0.78} & 59.50{\tiny$\pm$1.89} & \underline{67.84{\tiny$\pm$1.27}} & 69.60{\tiny$\pm$1.22} & 33.64{\tiny$\pm$0.77} & 64.58{\tiny$\pm$0.32} \\
    &\emph{NCNC} & 60.42{\tiny$\pm$1.89} & 83.22{\tiny$\pm$0.82} & 73.11{\tiny$\pm$2.07} & 57.13{\tiny$\pm$0.66} & 65.73{\tiny$\pm$2.57} & 72.87{\tiny$\pm$1.80} & 37.17{\tiny$\pm$1.86} & \textbf{65.97{\tiny$\pm$1.03}} \\
    
    \midrule
    \multirow{4}{*}{DAs}
    &\emph{Edge Proposal} & 70.19{\tiny$\pm$2.95} & 86.35{\tiny$\pm$1.35} & 81.59{\tiny$\pm$0.51} & 36.20{\tiny$\pm$2.61} & 62.44{\tiny$\pm$2.68} & 70.34{\tiny$\pm$2.89} & 33.76{\tiny$\pm$2.08} & 65.48{\tiny$\pm$0.00} \\
    &\emph{CFLP} & 54.36{\tiny$\pm$3.41} & 89.09{\tiny$\pm$1.12} & 73.57{\tiny$\pm$1.06} & 50.62{\tiny$\pm$3.33} & OOM & OOM & OOM & OOM \\
    &\emph{Node Drop} & 68.76{\tiny$\pm$2.77} & 90.79{\tiny$\pm$1.40} & 81.45{\tiny$\pm$3.10} & 61.76{\tiny$\pm$5.72} & 64.80{\tiny$\pm$2.52} & 70.51{\tiny$\pm$1.87} & 35.94{\tiny$\pm$2.30} & 62.57{\tiny$\pm$0.96} \\
    &\emph{Edge Drop} & 66.92{\tiny$\pm$4.29} & 92.12{\tiny$\pm$0.96} & 81.92{\tiny$\pm$1.94} & 59.66{\tiny$\pm$7.18} & 67.27{\tiny$\pm$1.64} & 72.52{\tiny$\pm$1.88} & 36.91{\tiny$\pm$0.94} & 63.20{\tiny$\pm$0.88} \\
    
    \midrule
    \multirow{3}{*}{Ours}
    &\emph{Complete Only} & \underline{72.10{\tiny$\pm$1.70}} & 91.84{\tiny$\pm$1.23} & 82.70{\tiny$\pm$2.20} & 63.18{\tiny$\pm$4.01} & 67.06{\tiny$\pm$1.01} & 71.83{\tiny$\pm$1.44} & 35.80{\tiny$\pm$1.38} & 63.57{\tiny$\pm$0.48} \\
    &\emph{Reduce Only} & 70.22{\tiny$\pm$3.69} & \underline{92.35{\tiny$\pm$0.95}} & \underline{84.22{\tiny$\pm$1.58}} & \underline{65.40{\tiny$\pm$2.27}} & 67.79{\tiny$\pm$1.50} & \underline{74.73{\tiny$\pm$2.12}} & \underline{37.88{\tiny$\pm$1.10}} & 64.24{\tiny$\pm$0.60} \\
    &\emph{CORE} & \textbf{76.34{\tiny$\pm$1.65}} & \textbf{93.14{\tiny$\pm$1.09}} & \textbf{84.67{\tiny$\pm$1.13}} & \textbf{65.64{\tiny$\pm$1.28}} & \textbf{69.67{\tiny$\pm$1.36}} & \textbf{74.73{\tiny$\pm$2.12}} & \textbf{37.88{\tiny$\pm$1.10}} & \underline{65.62{\tiny$\pm$0.50}} \\

    \midrule
    \multicolumn{2}{c}{\emph{p-values}} & 0.0001** & 0.0394** & 0.0096** & 0.0105** & 0.0060** & 0.0486** & 0.3126 & - \\
  \bottomrule
\end{NiceTabular}
}
\end{center}
\end{table*}


\begin{figure*}
\begin{center}
\centerline{\includegraphics[width=\textwidth]{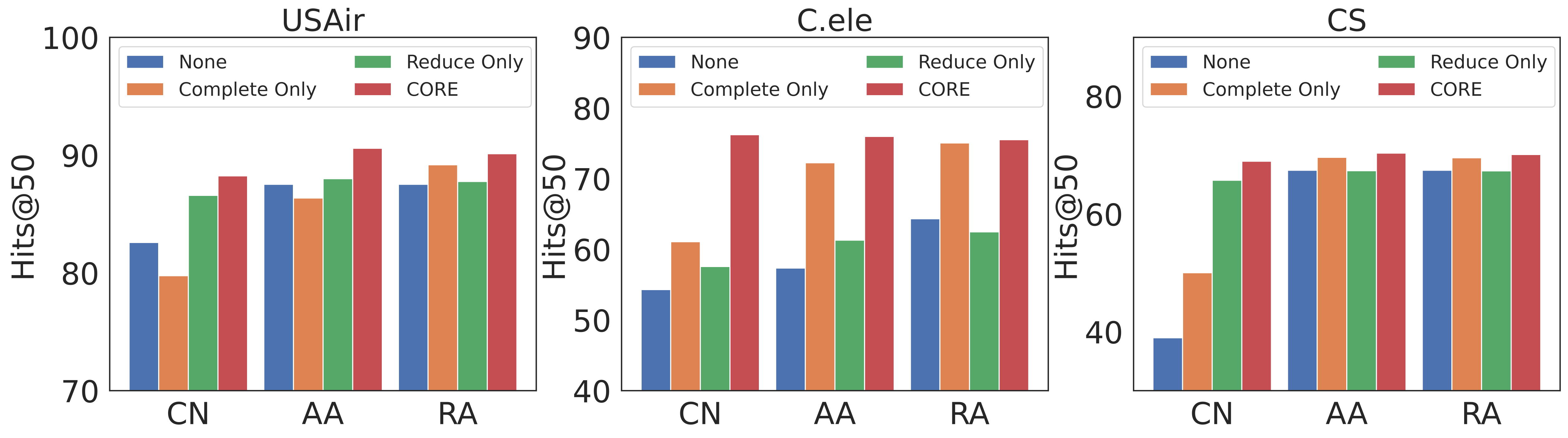}}
\caption{\our can enhance the graph structure and even boost heuristics link predictors (Hits@50).}\label{fig:enhanced}
\end{center}
\vspace{-15pt}
\end{figure*}

\begin{table}[h]
    \centering
\caption{Results 
of adversarial robustness for different models on \emph{C.ele} and \emph{USAir} datasets. 
The attack rates of 10\%, 30\%, and 50\% represent the respective ratios of edges subjected to adversarial flips by CLGA~\cite{zhang_unsupervised_2022}.}\label{table:robust}
\resizebox{1\linewidth}{!}{
\begin{tabular}{ll|ccccc}
    \toprule
    \textbf{Datasets} & \textbf{Methods} & \textbf{No Adv} & \textbf{10\%} & \textbf{30\%} & \textbf{50\%}\\
    
    \midrule
    \multirow{6}{*}{\textbf{C.ele}}
    &\emph{GCN} & 57.32{\tiny$\pm$4.52} & 59.63{\tiny$\pm$3.41} & 54.97{\tiny$\pm$3.08} & 46.76{\tiny$\pm$3.90} \\
    &\emph{SAGE} & 42.14{\tiny$\pm$5.62} & 31.98{\tiny$\pm$6.26} & 35.15{\tiny$\pm$3.38} & 28.32{\tiny$\pm$5.74} \\
    &\emph{SEAL} & 67.32{\tiny$\pm$2.71} & 60.93{\tiny$\pm$2.23} & 58.55{\tiny$\pm$1.46} & \underline{51.00{\tiny$\pm$2.32}} \\
    &\emph{ELPH} & 66.06{\tiny$\pm$3.00} & 62.28{\tiny$\pm$2.48} & 56.62{\tiny$\pm$2.48} & 50.40{\tiny$\pm$0.85} \\
    &\emph{NCNC} & 60.42{\tiny$\pm$1.89} & 52.10{\tiny$\pm$1.29} & 53.40{\tiny$\pm$1.40} & 50.26{\tiny$\pm$0.93}\\
    &\emph{Edge Proposal} & \underline{70.19{\tiny$\pm$2.95}} & \underline{64.71{\tiny$\pm$1.86}} & \underline{58.60{\tiny$\pm$2.14}} & 50.93{\tiny$\pm$2.00} \\
    &\emph{CFLP} & 54.36{\tiny$\pm$3.41} & 51.75{\tiny$\pm$2.79} & 46.49{\tiny$\pm$3.30} & 42.83{\tiny$\pm$5.16} \\
    &\emph{CORE} & \textbf{76.34{\tiny$\pm$1.65}} & \textbf{72.03{\tiny$\pm$3.19}} & \textbf{63.78{\tiny$\pm$2.24}} & \textbf{58.16{\tiny$\pm$1.52}} \\
    
    \midrule
    \multirow{6}{*}{\textbf{USAir}}
    &\emph{GCN} & 82.14{\tiny$\pm$1.99} & 84.87{\tiny$\pm$1.22} & 83.06{\tiny$\pm$1.73} & 80.19{\tiny$\pm$0.77} \\
    &\emph{SAGE} & 82.85{\tiny$\pm$4.01} & 78.21{\tiny$\pm$2.81} & 74.82{\tiny$\pm$3.28} & 73.88{\tiny$\pm$3.65} \\
    &\emph{SEAL} & \underline{91.76{\tiny$\pm$1.17}} & 85.51{\tiny$\pm$1.70} & 84.80{\tiny$\pm$2.95} & 81.53{\tiny$\pm$3.95} \\
    &\emph{ELPH} & 88.16{\tiny$\pm$1.21} & \underline{86.71{\tiny$\pm$0.94}} & \underline{85.08{\tiny$\pm$0.96}} & \underline{84.54{\tiny$\pm$0.50}}\\
    &\emph{NCNC} & 83.22{\tiny$\pm$0.82} & 83.88{\tiny$\pm$0.78} & 83.44{\tiny$\pm$0.50} & 83.18{\tiny$\pm$0.53}\\
    &\emph{Edge Proposal} & 86.35{\tiny$\pm$1.35} & 86.42{\tiny$\pm$1.34} & 84.95{\tiny$\pm$0.74} & 80.94{\tiny$\pm$1.66} \\
    &\emph{CFLP} & 89.09{\tiny$\pm$1.12} & 86.53{\tiny$\pm$1.74} & 77.26{\tiny$\pm$4.24} & 80.32{\tiny$\pm$2.44} \\
    &\emph{CORE} & \textbf{92.69{\tiny$\pm$0.75}} & \textbf{89.72{\tiny$\pm$1.06}} & \textbf{88.02{\tiny$\pm$1.13}} & \textbf{86.71{\tiny$\pm$2.06}} \\
  \bottomrule
\end{tabular}
}
\end{table}

\begin{table}[h]
    \centering
\caption{Ablation study. 
 The upper half of the table presents results for~\our with GIN as the backbone model, while the bottom half investigates the impact of the balancing hyperparameter $\beta$ and edge sampling in the proposed framework.}\label{table:ablation}
 \resizebox{1\linewidth}{!}{
\begin{tabular}{l cccc}
    \toprule
    \textbf{Methods} & \textbf{C.ele} & \textbf{USAir} & \textbf{Router} & \textbf{Yeast}\\
    
    \midrule
    \multicolumn{5}{l}{\emph{GIN} as the backbone model} \\
    \emph{GIN} & 62.77{\tiny$\pm$2.33} & 87.22{\tiny$\pm$2.70} & \underline{60.22{\tiny$\pm$2.09}} & \underline{75.38{\tiny$\pm$2.23}} \\
    \emph{Complete Only} & \underline{71.03{\tiny$\pm$2.18}} & \underline{88.12{\tiny$\pm$1.47}} & 60.22{\tiny$\pm$2.09} & 75.38{\tiny$\pm$2.23} \\
    \emph{Reduce Only} & 64.13{\tiny$\pm$2.84} & 88.71{\tiny$\pm$1.60} & 62.81{\tiny$\pm$2.46} & 78.40{\tiny$\pm$1.34} \\ 
    \emph{CORE} & \textbf{72.33{\tiny$\pm$2.62}} & \textbf{88.71{\tiny$\pm$1.60}} & \textbf{62.81{\tiny$\pm$2.46}} & \textbf{78.40{\tiny$\pm$1.34}} \\
    \midrule
    \multicolumn{5}{l}{\emph{CORE} without sampling or info constraint} \\
    \emph{NoSample} & \underline{75.20{\tiny$\pm$1.71}} & 91.51{\tiny$\pm$1.65} & 64.35{\tiny$\pm$2.49} & \textbf{84.59{\tiny$\pm$1.16}} \\
    \emph{$\beta=0$} & 74.02{\tiny$\pm$2.45} & \underline{91.81{\tiny$\pm$1.53}} & 63.90{\tiny$\pm$2.16} & 83.33{\tiny$\pm$2.05} \\
    \emph{NoSample-$\beta=0$} & 73.48{\tiny$\pm$2.52} & 91.45{\tiny$\pm$1.73} & \underline{65.09{\tiny$\pm$1.41}} & \underline{84.47{\tiny$\pm$1.49}} \\ 
    \emph{CORE} & \textbf{76.34{\tiny$\pm$1.65}} & \textbf{92.69{\tiny$\pm$0.75}} & \textbf{65.47{\tiny$\pm$2.44}} & 84.22{\tiny$\pm$1.58} \\
  \bottomrule
\end{tabular}
}
\end{table}

In this section, we present experimental results for our proposed method. 
We first assess the performance of~\our in comparison to various 
baseline DA techniques for the link prediction task.
Then, we illustrate that heuristic link predictors can also benefit from the augmented graph structure by~\our.
Furthermore, we demonstrate its robustness against adversarial edge perturbations. 
Further details of the experiments can be found in Appendix \ref{app:implement}.

\subsection{Experimental setup}

\paragraph{Baseline methods.}
We select three heuristic link predictors for non-GNN models: 
\emph{CN}~\cite{newman_finding_2006}, \emph{AA}~\cite{adamic_friends_2003}, and \emph{RA}~\cite{zhou_predicting_2009}.
For GNN models that exploit node-level representation, we employ the two most widely used architectures: 
\emph{GCN}~\cite{kipf_semi-supervised_2017} and \emph{SAGE}~\cite{hamilton_inductive_2018}. 
For the link prediction utilizing edge-level representation,
we choose \emph{SEAL}~\cite{zhang_link_2018}, \emph{ELPH}~\cite{chamberlain_graph_2022} and \emph{NCNC}~\cite{wang_neural_2023} as the baseline.

We select \emph{Edge Proposal}~\cite{singh_edge_2021} and \emph{CFLP}~\cite{zhao_learning_2021}, 
as two representative DA baselines with \emph{GCN} as backbone.
For \emph{SEAL},
we evaluate two standard graph perturbation techniques~\cite{you_graph_2021},~\emph{Node Drop}~\cite{papp_dropgnn_2021} and 
\emph{Edge Drop}~\cite{rong_dropedge_2020}.
Then, we present \textbf{our} results with~\emph{Complete Only},~\emph{Reduce Only}, and the combined~\emph{\our}. 
Details about these baseline models can be found in Appendix \ref{app:baseline}.
\vspace{-5pt}
\paragraph{Benchmark datasets.} 
We select four attributed and four non-attributed graphs as the benchmark. The attributed graphs consist of three collaboration networks, 
\textbf{CS}, \textbf{Physics}~\cite{shchur_pitfalls_2019} and \textbf{Collab}~\cite{wu_graph_2020}, as well as a co-purchased graph, \textbf{Computers}~\cite{shchur_pitfalls_2019}.
The non-attributed graphs include \textbf{USAir}~\cite{batagelj_pajek_2006}, \textbf{Yeast}~\cite{von_mering_comparative_2002}, \textbf{C.ele}~\cite{watts_collective_1998},
and \textbf{Router}~\cite{spring_measuring_2002}. The comprehensive descriptions and statistics of the benchmark datasets can be found in Appendix \ref{app:benchmark_data}.
\vspace{-5pt}
\paragraph{Evaluation protocols.}
We follow the evaluation settings from previous work~\cite{zhao_learning_2021} and
split the links as $10\%$ for validation, $20\%$ for testing.
For~\textbf{Collab}~\cite{hu_open_2021}, we use the official train-test split.
The evaluation metric is Hits@50, which is widely accepted for evaluating link prediction tasks~\cite{hu_open_2021}. 
The results are reported for 10 different runs with varying model initializations.

\subsection{Experimental results}
\paragraph{Link prediction.}
\autoref{table:main} presents the link prediction performance of Hits@50
for all methods. Given the strong backbone model \emph{SEAL},
we observe that our proposed data augmentation can further improve
its performance on various datasets. In comparison to \emph{SEAL} without any
DA techniques, \our consistently boosts the performance
by $1\%$ to $9\%$ in terms of Hits@50. More specifically, both
\emph{Complete Only} and \emph{Reduce Only} can increase the model
capability by different margins.
Moreover, by combining those two stages together,
\our can almost always achieve the best performance 
and significantly outperforms baselines.
Our results also reveal that~\our yields greater performance improvements when the available data size is limited. 
This observation suggests that models may be prone to overfitting to noise in low-data regimes. 
However,~\our effectively mitigates this issue by learning an underlying (Bernoulli) distribution associated with the graph structures, 
and prevents the model from overfitting to idiosyncratic structural perturbations.
\vspace{-5pt}
\paragraph{Learnable and transferrable.}
One potential concern with using the reducer of~\our, 
which is a neural network possessing 
the capability of universal approximation~\cite{hornik_multilayer_1989}, 
is that the performance improvement might be attributed to the overparameterization~\cite{belkin_fit_2021} of the model
instead of the quality of our augmented graph.
To address this concern and validate the efficacy of~\our as a DA method, we decouple the reducer from the model and investigate its ability to extract a generalizable view of the graph.
We feed the augmented graph generated by the reducer to 
three heuristic link predictors: \emph{CN}, \emph{AA}, and \emph{RA}. 
The results of this experiment can be found in~\Figref{fig:enhanced}. 
Our findings demonstrate that the graph refined by~\our consistently 
improves the performance of heuristic link predictors. 
This outcome validates ~\our's ability to learn a transferable and generalizable DA. 
\kevin{the heuristics method for weighted graph in appendix}
\vspace{-5pt}
\paragraph{Robustness.}
To assess the robustness of our graph data augmentation method, 
we conduct additional experiments using an unsupervised graph poisoning attack, CLGA~\cite{zhang_unsupervised_2022}, to adversarially perturb the graph structures at varying attack rates. 
The results of this analysis can be found in Table~\ref{table:robust} and Table~\ref{table:robust2}. 
Intriguingly, we observe that the advanced link prediction models, like \emph{SEAL}, \emph{ELPH} and \emph{NCNC}, exhibits a higher vulnerability to adversarial attacks compared to other baseline models. 
The capability to capture complex structural relationships of these expressive models renders them more sensitive to structural changes.
However, our proposed method,~\our, leverages the robustness inherent in IB~\cite{alemi_deep_2023,wu_graph_2020} 
to enhance the model resilience by pruning spurious or even harmful perturbations. 
These findings suggest that the performance improvement offered by our method 
may be attributed to its ability to mitigate model vulnerability to adversarial perturbations.
\vspace{-5pt}
\paragraph{\emph{GIN} as backbone.}
We examine whether~\our remains an effective DA technique when utilizing a different backbone model. 
In this case, we choose the Graph Isomorphism Network (\emph{GIN})~\cite{xu_how_2018}, one of the most expressive GNNs, ensuring that the learned representation can encode structural information 
and guide downstream data augmentation. 
The results are presented in upper half of Table~\ref{table:ablation}. 
We observe that, with \emph{GIN} as the backbone,~\our can still improve link prediction performance over the baseline, yielding a $1\%$ to $10\%$ improvement.
It indicates that~\our can be effectively integrated with other backbone models.
\kevin{Appendix: zero one node labelling}
\paragraph{Information constraints and stochastic sampling.}
We further investigate the necessity of retaining the information constraint term in the objective function and the stochastic sampling component in the augmentation process. 
The results are displayed in the lower half of Table~\ref{table:ablation}. By setting $\beta=0$, the Reduce stage of our method loses the ability to constrain the information flow from the inflated graph. 
This leads to significant performance degradation, suggesting that the regularization term helps prevent the model from overfitting. 
Additionally, the performance declines when removing the stochastic sampling component and directly applying the attention score as the edge weight. 
This demonstrates that incorporating sampling in data augmentation can potentially expose the link prediction model to a wider range of augmented data variations.

\subsection{Different DAs for different target links}
\label{sec:edge_dis}
\begin{figure*}[t]
\begin{center}
\centerline{\includegraphics[width=\textwidth]{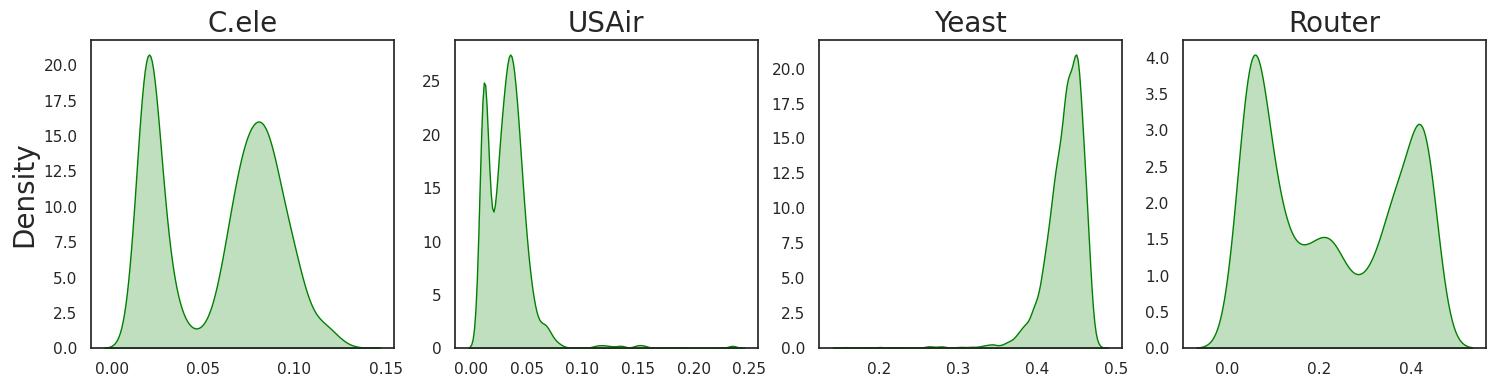}}
\caption{Histogram representing the standard deviations (std) of the learned edge mask $\omega$ for each edge within subgraphs associated with different target links. The frequent occurrence of larger std values implies substantial disagreement on the optimal DAs when focusing on different target links.}
\label{fig:edge_dis}
\end{center}
\vspace{-15pt}
\end{figure*}

One of the unique designs of our methods is to augment each target link in a separate environment of its own. Here, we empirically investigate the necessity of isolating the DAs. We collect the edge mask $\omega$ for each edge within the subgraphs but across different target links. Then, for a set of such edge masks of the same edge, we calculate their standard deviations to indicate how much the learned edge masks $\omega$ agree or disagree with each other when augmenting different target links. The results are presented in~\Figref{fig:edge_dis}. 

As it shows, while a portion of edges may have similar augmentation (small standard deviations), a significant part of them conflicts with each other (large standard deviations). On~\textbf{C.ele} and~\textbf{Router},~\our will learn different DAs for nearly half of the edges associated with different target links. On~\textbf{Yeast}, the majority of edges are augmented differently by~\our. This result indicates that it is necessary to isolate the DA effect for each target link.

\subsection{Additional ablation studies}
To further substantiate the efficacy of our proposed DA method, we carry out extensive ablation studies. Due to page constraints, these detailed investigations are presented in the appendix. They include an analysis on the impact of different components of the Reduce stage (see Appendix \ref{app:ab}), a study on parameter sensitivity (see Appendix \ref{app:param_sen}), and evaluations of \our when integrated with GCN and SAGE backbones (see Appendix \ref{app:gcn_sage}).
\vspace{-3mm}
\section{Conclusion}
In this paper, we have introduced~\our, 
a novel data augmentation technique specifically designed for link prediction tasks. 
Leveraging the Information Bottleneck principle,
\our effectively eliminates noisy and spurious edges while recovering missing edges in the graph, 
thereby enhancing the generalizability of link prediction models. 
Our approach yields graph structures that reveal the fundamental relationships inherent in the graph. 
Extensive experiments on various benchmark datasets have demonstrated the effectiveness and superiority of~\our over competing methods,
highlighting its potential as a leading approach for robust link prediction in graph representation learning.

\clearpage
\bibliographystyle{ACM-Reference-Format}
\balance
\bibliography{references}


\begin{thebibliography}{72}


\ifx \showCODEN    \undefined \def \showCODEN     #1{\unskip}     \fi
\ifx \showDOI      \undefined \def \showDOI       #1{#1}\fi
\ifx \showISBNx    \undefined \def \showISBNx     #1{\unskip}     \fi
\ifx \showISBNxiii \undefined \def \showISBNxiii  #1{\unskip}     \fi
\ifx \showISSN     \undefined \def \showISSN      #1{\unskip}     \fi
\ifx \showLCCN     \undefined \def \showLCCN      #1{\unskip}     \fi
\ifx \shownote     \undefined \def \shownote      #1{#1}          \fi
\ifx \showarticletitle \undefined \def \showarticletitle #1{#1}   \fi
\ifx \showURL      \undefined \def \showURL       {\relax}        \fi
\providecommand\bibfield[2]{#2}
\providecommand\bibinfo[2]{#2}
\providecommand\natexlab[1]{#1}
\providecommand\showeprint[2][]{arXiv:#2}

\bibitem[Adamic and Adar(2003)]%
        {adamic_friends_2003}
\bibfield{author}{\bibinfo{person}{Lada~A. Adamic} {and} \bibinfo{person}{Eytan
  Adar}.} \bibinfo{year}{2003}\natexlab{}.
\newblock \showarticletitle{Friends and neighbors on the {Web}}.
\newblock \bibinfo{journal}{\emph{Social Networks}} \bibinfo{volume}{25},
  \bibinfo{number}{3} (\bibinfo{year}{2003}), \bibinfo{pages}{211--230}.
\newblock
\showISSN{0378-8733}
\urldef\tempurl%
\url{https://doi.org/10.1016/S0378-8733(03)00009-1}
\showDOI{\tempurl}


\bibitem[Alemi et~al\mbox{.}(2023)]%
        {alemi_deep_2023}
\bibfield{author}{\bibinfo{person}{Alexander~A. Alemi}, \bibinfo{person}{Ian
  Fischer}, \bibinfo{person}{Joshua~V. Dillon}, {and} \bibinfo{person}{Kevin
  Murphy}.} \bibinfo{year}{2023}\natexlab{}.
\newblock \showarticletitle{Deep {Variational} {Information} {Bottleneck}}.
\newblock
\urldef\tempurl%
\url{https://openreview.net/forum?id=HyxQzBceg}
\showURL{%
\tempurl}


\bibitem[Arjovsky et~al\mbox{.}(2019)]%
        {arjovsky_invariant_2019}
\bibfield{author}{\bibinfo{person}{Martin Arjovsky}, \bibinfo{person}{Léon
  Bottou}, \bibinfo{person}{Ishaan Gulrajani}, {and} \bibinfo{person}{David
  Lopez-Paz}.} \bibinfo{year}{2019}\natexlab{}.
\newblock \bibinfo{title}{Invariant {Risk} {Minimization}}.
\newblock
\newblock
\urldef\tempurl%
\url{https://arxiv.org/abs/1907.02893v3}
\showURL{%
\tempurl}


\bibitem[Batagelj and Mrvar(2006)]%
        {batagelj_pajek_2006}
\bibfield{author}{\bibinfo{person}{Vladimir Batagelj} {and}
  \bibinfo{person}{Andrej Mrvar}.} \bibinfo{year}{2006}\natexlab{}.
\newblock \bibinfo{title}{Pajek datasets website}.
\newblock
\newblock
\urldef\tempurl%
\url{http://vlado.fmf.uni-lj.si/pub/networks/data/}
\showURL{%
\tempurl}


\bibitem[Belkin(2021)]%
        {belkin_fit_2021}
\bibfield{author}{\bibinfo{person}{Mikhail Belkin}.}
  \bibinfo{year}{2021}\natexlab{}.
\newblock \bibinfo{title}{Fit without fear: remarkable mathematical phenomena
  of deep learning through the prism of interpolation}.
\newblock
\newblock
\urldef\tempurl%
\url{https://doi.org/10.48550/arXiv.2105.14368}
\showDOI{\tempurl}
\newblock
\shownote{arXiv:2105.14368 [cs, math, stat]}.


\bibitem[Brin and Page(1998)]%
        {brin_anatomy_1998}
\bibfield{author}{\bibinfo{person}{Sergey Brin} {and} \bibinfo{person}{Lawrence
  Page}.} \bibinfo{year}{1998}\natexlab{}.
\newblock \showarticletitle{The {Anatomy} of a {Large}-{Scale} {Hypertextual}
  {Web} {Search} {Engine}}.
\newblock \bibinfo{journal}{\emph{Computer Networks}}  \bibinfo{volume}{30}
  (\bibinfo{year}{1998}), \bibinfo{pages}{107--117}.
\newblock
\urldef\tempurl%
\url{http://www-db.stanford.edu/~backrub/google.html}
\showURL{%
\tempurl}


\bibitem[Chamberlain et~al\mbox{.}(2022)]%
        {chamberlain_graph_2022}
\bibfield{author}{\bibinfo{person}{Benjamin~Paul Chamberlain},
  \bibinfo{person}{Sergey Shirobokov}, \bibinfo{person}{Emanuele Rossi},
  \bibinfo{person}{Fabrizio Frasca}, \bibinfo{person}{Thomas Markovich},
  \bibinfo{person}{Nils~Yannick Hammerla}, \bibinfo{person}{Michael~M.
  Bronstein}, {and} \bibinfo{person}{Max Hansmire}.}
  \bibinfo{year}{2022}\natexlab{}.
\newblock \showarticletitle{Graph {Neural} {Networks} for {Link} {Prediction}
  with {Subgraph} {Sketching}}.
\newblock
\urldef\tempurl%
\url{https://openreview.net/forum?id=m1oqEOAozQU}
\showURL{%
\tempurl}


\bibitem[Chawla et~al\mbox{.}(2002)]%
        {chawla_smote_2002}
\bibfield{author}{\bibinfo{person}{N.~V. Chawla}, \bibinfo{person}{K.~W.
  Bowyer}, \bibinfo{person}{L.~O. Hall}, {and} \bibinfo{person}{W.~P.
  Kegelmeyer}.} \bibinfo{year}{2002}\natexlab{}.
\newblock \showarticletitle{{SMOTE}: {Synthetic} {Minority} {Over}-sampling
  {Technique}}.
\newblock \bibinfo{journal}{\emph{Journal of Artificial Intelligence Research}}
   \bibinfo{volume}{16} (\bibinfo{date}{June} \bibinfo{year}{2002}),
  \bibinfo{pages}{321--357}.
\newblock
\showISSN{1076-9757}
\urldef\tempurl%
\url{https://doi.org/10.1613/jair.953}
\showDOI{\tempurl}
\newblock
\shownote{arXiv:1106.1813 [cs]}.


\bibitem[Chen et~al\mbox{.}(2020)]%
        {chen_iterative_2020}
\bibfield{author}{\bibinfo{person}{Yu Chen}, \bibinfo{person}{Lingfei Wu},
  {and} \bibinfo{person}{Mohammed Zaki}.} \bibinfo{year}{2020}\natexlab{}.
\newblock \showarticletitle{Iterative {Deep} {Graph} {Learning} for {Graph}
  {Neural} {Networks}: {Better} and {Robust} {Node} {Embeddings}}. In
  \bibinfo{booktitle}{\emph{Advances in {Neural} {Information} {Processing}
  {Systems}}}, Vol.~\bibinfo{volume}{33}. \bibinfo{publisher}{Curran
  Associates, Inc.}, \bibinfo{pages}{19314--19326}.
\newblock
\urldef\tempurl%
\url{https://proceedings.neurips.cc/paper/2020/hash/e05c7ba4e087beea9410929698dc41a6-Abstract.html}
\showURL{%
\tempurl}


\bibitem[Cubuk et~al\mbox{.}(2019)]%
        {cubuk_autoaugment_2019}
\bibfield{author}{\bibinfo{person}{Ekin~D. Cubuk}, \bibinfo{person}{Barret
  Zoph}, \bibinfo{person}{Dandelion Mane}, \bibinfo{person}{Vijay Vasudevan},
  {and} \bibinfo{person}{Quoc~V. Le}.} \bibinfo{year}{2019}\natexlab{}.
\newblock \bibinfo{title}{{AutoAugment}: {Learning} {Augmentation} {Policies}
  from {Data}}.
\newblock
\newblock
\urldef\tempurl%
\url{https://doi.org/10.48550/arXiv.1805.09501}
\showDOI{\tempurl}
\newblock
\shownote{arXiv:1805.09501 [cs, stat]}.


\bibitem[DeVries and Taylor(2017)]%
        {devries_improved_2017}
\bibfield{author}{\bibinfo{person}{Terrance DeVries} {and}
  \bibinfo{person}{Graham~W. Taylor}.} \bibinfo{year}{2017}\natexlab{}.
\newblock \bibinfo{title}{Improved {Regularization} of {Convolutional} {Neural}
  {Networks} with {Cutout}}.
\newblock
\newblock
\urldef\tempurl%
\url{https://doi.org/10.48550/arXiv.1708.04552}
\showDOI{\tempurl}
\newblock
\shownote{arXiv:1708.04552 [cs]}.


\bibitem[Dong et~al\mbox{.}(2022)]%
        {dong_fakeedge_2022}
\bibfield{author}{\bibinfo{person}{Kaiwen Dong}, \bibinfo{person}{Yijun Tian},
  \bibinfo{person}{Zhichun Guo}, \bibinfo{person}{Yang Yang}, {and}
  \bibinfo{person}{Nitesh Chawla}.} \bibinfo{year}{2022}\natexlab{}.
\newblock \showarticletitle{{FakeEdge}: {Alleviate} {Dataset} {Shift} in {Link}
  {Prediction}}.
\newblock
\urldef\tempurl%
\url{https://openreview.net/forum?id=QDN0jSXuvtX}
\showURL{%
\tempurl}


\bibitem[Erdos(1959)]%
        {erdos_random_1959}
\bibfield{author}{\bibinfo{person}{Paul Erdos}.}
  \bibinfo{year}{1959}\natexlab{}.
\newblock \showarticletitle{On random graphs}.
\newblock \bibinfo{journal}{\emph{Publicationes mathematicae}}
  \bibinfo{volume}{6} (\bibinfo{year}{1959}), \bibinfo{pages}{290--297}.
\newblock


\bibitem[Feng et~al\mbox{.}(2021)]%
        {feng_survey_2021}
\bibfield{author}{\bibinfo{person}{Steven~Y. Feng}, \bibinfo{person}{Varun
  Gangal}, \bibinfo{person}{Jason Wei}, \bibinfo{person}{Sarath Chandar},
  \bibinfo{person}{Soroush Vosoughi}, \bibinfo{person}{Teruko Mitamura}, {and}
  \bibinfo{person}{Eduard Hovy}.} \bibinfo{year}{2021}\natexlab{}.
\newblock \showarticletitle{A {Survey} of {Data} {Augmentation} {Approaches}
  for {NLP}}. In \bibinfo{booktitle}{\emph{Findings of the {Association} for
  {Computational} {Linguistics}: {ACL}-{IJCNLP} 2021}}.
  \bibinfo{publisher}{Association for Computational Linguistics},
  \bibinfo{address}{Online}, \bibinfo{pages}{968--988}.
\newblock
\urldef\tempurl%
\url{https://doi.org/10.18653/v1/2021.findings-acl.84}
\showDOI{\tempurl}


\bibitem[Fey and Lenssen(2019)]%
        {fey_fast_2019}
\bibfield{author}{\bibinfo{person}{Matthias Fey} {and} \bibinfo{person}{Jan~E.
  Lenssen}.} \bibinfo{year}{2019}\natexlab{}.
\newblock \showarticletitle{Fast {Graph} {Representation} {Learning} with
  {PyTorch} {Geometric}}. In \bibinfo{booktitle}{\emph{{ICLR} {Workshop} on
  {Representation} {Learning} on {Graphs} and {Manifolds}}}.
\newblock


\bibitem[Gilmer et~al\mbox{.}(2017)]%
        {gilmer_neural_2017}
\bibfield{author}{\bibinfo{person}{Justin Gilmer}, \bibinfo{person}{Samuel~S.
  Schoenholz}, \bibinfo{person}{Patrick~F. Riley}, \bibinfo{person}{Oriol
  Vinyals}, {and} \bibinfo{person}{George~E. Dahl}.}
  \bibinfo{year}{2017}\natexlab{}.
\newblock \showarticletitle{Neural {Message} {Passing} for {Quantum}
  {Chemistry}}.
\newblock \bibinfo{journal}{\emph{CoRR}}  \bibinfo{volume}{abs/1704.01212}
  (\bibinfo{year}{2017}).
\newblock
\urldef\tempurl%
\url{http://arxiv.org/abs/1704.01212}
\showURL{%
\tempurl}
\newblock
\shownote{arXiv: 1704.01212}.


\bibitem[Hamilton et~al\mbox{.}(2018)]%
        {hamilton_inductive_2018}
\bibfield{author}{\bibinfo{person}{William~L. Hamilton}, \bibinfo{person}{Rex
  Ying}, {and} \bibinfo{person}{Jure Leskovec}.}
  \bibinfo{year}{2018}\natexlab{}.
\newblock \showarticletitle{Inductive {Representation} {Learning} on {Large}
  {Graphs}}.
\newblock \bibinfo{journal}{\emph{arXiv:1706.02216 [cs, stat]}}
  (\bibinfo{date}{Sept.} \bibinfo{year}{2018}).
\newblock
\urldef\tempurl%
\url{http://arxiv.org/abs/1706.02216}
\showURL{%
\tempurl}
\newblock
\shownote{arXiv: 1706.02216}.


\bibitem[Holland et~al\mbox{.}(1983)]%
        {holland_stochastic_1983}
\bibfield{author}{\bibinfo{person}{Paul Holland}, \bibinfo{person}{Kathryn~B.
  Laskey}, {and} \bibinfo{person}{Samuel Leinhardt}.}
  \bibinfo{year}{1983}\natexlab{}.
\newblock \showarticletitle{Stochastic blockmodels: {First} steps}.
\newblock \bibinfo{journal}{\emph{Social Networks}}  \bibinfo{volume}{5}
  (\bibinfo{year}{1983}), \bibinfo{pages}{109--137}.
\newblock
\urldef\tempurl%
\url{https://api.semanticscholar.org/CorpusID:34098453}
\showURL{%
\tempurl}


\bibitem[Hornik et~al\mbox{.}(1989)]%
        {hornik_multilayer_1989}
\bibfield{author}{\bibinfo{person}{Kurt Hornik}, \bibinfo{person}{Maxwell
  Stinchcombe}, {and} \bibinfo{person}{Halbert White}.}
  \bibinfo{year}{1989}\natexlab{}.
\newblock \showarticletitle{Multilayer feedforward networks are universal
  approximators}.
\newblock \bibinfo{journal}{\emph{Neural Networks}} \bibinfo{volume}{2},
  \bibinfo{number}{5} (\bibinfo{date}{Jan.} \bibinfo{year}{1989}),
  \bibinfo{pages}{359--366}.
\newblock
\showISSN{0893-6080}
\urldef\tempurl%
\url{https://doi.org/10.1016/0893-6080(89)90020-8}
\showDOI{\tempurl}


\bibitem[Hu et~al\mbox{.}(2021)]%
        {hu_open_2021}
\bibfield{author}{\bibinfo{person}{Weihua Hu}, \bibinfo{person}{Matthias Fey},
  \bibinfo{person}{Marinka Zitnik}, \bibinfo{person}{Yuxiao Dong},
  \bibinfo{person}{Hongyu Ren}, \bibinfo{person}{Bowen Liu},
  \bibinfo{person}{Michele Catasta}, {and} \bibinfo{person}{Jure Leskovec}.}
  \bibinfo{year}{2021}\natexlab{}.
\newblock \showarticletitle{Open {Graph} {Benchmark}: {Datasets} for {Machine}
  {Learning} on {Graphs}}.
\newblock \bibinfo{journal}{\emph{arXiv:2005.00687 [cs, stat]}}
  (\bibinfo{date}{Feb.} \bibinfo{year}{2021}).
\newblock
\urldef\tempurl%
\url{http://arxiv.org/abs/2005.00687}
\showURL{%
\tempurl}
\newblock
\shownote{arXiv: 2005.00687}.


\bibitem[Hwang et~al\mbox{.}(2022)]%
        {hwang_analysis_2022}
\bibfield{author}{\bibinfo{person}{EunJeong Hwang}, \bibinfo{person}{Veronika
  Thost}, \bibinfo{person}{Shib~Sankar Dasgupta}, {and}
  \bibinfo{person}{Tengfei Ma}.} \bibinfo{year}{2022}\natexlab{}.
\newblock \showarticletitle{An {Analysis} of {Virtual} {Nodes} in {Graph}
  {Neural} {Networks} for {Link} {Prediction} ({Extended} {Abstract})}.
\newblock
\urldef\tempurl%
\url{https://openreview.net/forum?id=dI6KBKNRp7}
\showURL{%
\tempurl}


\bibitem[Jang et~al\mbox{.}(2023)]%
        {jang_categorical_2023}
\bibfield{author}{\bibinfo{person}{Eric Jang}, \bibinfo{person}{Shixiang Gu},
  {and} \bibinfo{person}{Ben Poole}.} \bibinfo{year}{2023}\natexlab{}.
\newblock \showarticletitle{Categorical {Reparameterization} with
  {Gumbel}-{Softmax}}.
\newblock
\urldef\tempurl%
\url{https://openreview.net/forum?id=rkE3y85ee}
\showURL{%
\tempurl}


\bibitem[Katz(1953)]%
        {katz_new_1953}
\bibfield{author}{\bibinfo{person}{Leo Katz}.} \bibinfo{year}{1953}\natexlab{}.
\newblock \showarticletitle{A new status index derived from sociometric
  analysis}.
\newblock \bibinfo{journal}{\emph{Psychometrika}} \bibinfo{volume}{18},
  \bibinfo{number}{1} (\bibinfo{date}{March} \bibinfo{year}{1953}),
  \bibinfo{pages}{39--43}.
\newblock
\showISSN{1860-0980}
\urldef\tempurl%
\url{https://doi.org/10.1007/BF02289026}
\showDOI{\tempurl}


\bibitem[Kingma and Welling(2014)]%
        {kingma_auto-encoding_2014}
\bibfield{author}{\bibinfo{person}{Diederik~P. Kingma} {and}
  \bibinfo{person}{Max Welling}.} \bibinfo{year}{2014}\natexlab{}.
\newblock \showarticletitle{Auto-{Encoding} {Variational} {Bayes}}.
\newblock \bibinfo{journal}{\emph{arXiv:1312.6114 [cs, stat]}}
  (\bibinfo{date}{May} \bibinfo{year}{2014}).
\newblock
\urldef\tempurl%
\url{http://arxiv.org/abs/1312.6114}
\showURL{%
\tempurl}
\newblock
\shownote{arXiv: 1312.6114}.


\bibitem[Kipf and Welling(2016)]%
        {kipf_variational_2016}
\bibfield{author}{\bibinfo{person}{Thomas~N. Kipf} {and} \bibinfo{person}{Max
  Welling}.} \bibinfo{year}{2016}\natexlab{}.
\newblock \bibinfo{title}{Variational {Graph} {Auto}-{Encoders}}.
\newblock
\newblock
\newblock
\shownote{\_eprint: 1611.07308}.


\bibitem[Kipf and Welling(2017)]%
        {kipf_semi-supervised_2017}
\bibfield{author}{\bibinfo{person}{Thomas~N. Kipf} {and} \bibinfo{person}{Max
  Welling}.} \bibinfo{year}{2017}\natexlab{}.
\newblock \showarticletitle{Semi-{Supervised} {Classification} with {Graph}
  {Convolutional} {Networks}}.
\newblock \bibinfo{journal}{\emph{arXiv:1609.02907 [cs, stat]}}
  (\bibinfo{date}{Feb.} \bibinfo{year}{2017}).
\newblock
\urldef\tempurl%
\url{http://arxiv.org/abs/1609.02907}
\showURL{%
\tempurl}
\newblock
\shownote{arXiv: 1609.02907}.


\bibitem[Koren et~al\mbox{.}(2009)]%
        {koren_matrix_2009}
\bibfield{author}{\bibinfo{person}{Yehuda Koren}, \bibinfo{person}{Robert
  Bell}, {and} \bibinfo{person}{Chris Volinsky}.}
  \bibinfo{year}{2009}\natexlab{}.
\newblock \showarticletitle{Matrix factorization techniques for recommender
  systems}.
\newblock \bibinfo{journal}{\emph{Computer}} \bibinfo{volume}{42},
  \bibinfo{number}{8} (\bibinfo{year}{2009}), \bibinfo{pages}{30--37}.
\newblock
\newblock
\shownote{Publisher: IEEE}.


\bibitem[Krueger et~al\mbox{.}(2020)]%
        {krueger_out--distribution_2020}
\bibfield{author}{\bibinfo{person}{David Krueger}, \bibinfo{person}{Ethan
  Caballero}, \bibinfo{person}{Joern-Henrik Jacobsen}, \bibinfo{person}{Amy
  Zhang}, \bibinfo{person}{Jonathan Binas}, \bibinfo{person}{Dinghuai Zhang},
  \bibinfo{person}{Remi~Le Priol}, {and} \bibinfo{person}{Aaron Courville}.}
  \bibinfo{year}{2020}\natexlab{}.
\newblock \bibinfo{title}{Out-of-{Distribution} {Generalization} via {Risk}
  {Extrapolation} ({REx})}.
\newblock
\newblock
\urldef\tempurl%
\url{https://arxiv.org/abs/2003.00688v5}
\showURL{%
\tempurl}


\bibitem[Li and Spratling(2023)]%
        {li_data_2023}
\bibfield{author}{\bibinfo{person}{Lin Li} {and} \bibinfo{person}{Michael
  Spratling}.} \bibinfo{year}{2023}\natexlab{}.
\newblock \bibinfo{title}{Data {Augmentation} {Alone} {Can} {Improve}
  {Adversarial} {Training}}.
\newblock
\newblock
\urldef\tempurl%
\url{https://arxiv.org/abs/2301.09879v1}
\showURL{%
\tempurl}


\bibitem[Li et~al\mbox{.}(2020)]%
        {li_distance_2020}
\bibfield{author}{\bibinfo{person}{Pan Li}, \bibinfo{person}{Yanbang Wang},
  \bibinfo{person}{Hongwei Wang}, {and} \bibinfo{person}{Jure Leskovec}.}
  \bibinfo{year}{2020}\natexlab{}.
\newblock \showarticletitle{Distance {Encoding}: {Design} {Provably} {More}
  {Powerful} {Neural} {Networks} for {Graph} {Representation} {Learning}}. In
  \bibinfo{booktitle}{\emph{Advances in {Neural} {Information} {Processing}
  {Systems}}}, \bibfield{editor}{\bibinfo{person}{H.~Larochelle},
  \bibinfo{person}{M.~Ranzato}, \bibinfo{person}{R.~Hadsell},
  \bibinfo{person}{M.~F. Balcan}, {and} \bibinfo{person}{H.~Lin}} (Eds.),
  Vol.~\bibinfo{volume}{33}. \bibinfo{publisher}{Curran Associates, Inc.},
  \bibinfo{pages}{4465--4478}.
\newblock
\urldef\tempurl%
\url{https://proceedings.neurips.cc/paper/2020/file/2f73168bf3656f697507752ec592c437-Paper.pdf}
\showURL{%
\tempurl}


\bibitem[Liben-Nowell and Kleinberg(2003)]%
        {liben-nowell_link_2003}
\bibfield{author}{\bibinfo{person}{David Liben-Nowell} {and}
  \bibinfo{person}{Jon Kleinberg}.} \bibinfo{year}{2003}\natexlab{}.
\newblock \showarticletitle{The link prediction problem for social networks}.
  In \bibinfo{booktitle}{\emph{Proceedings of the twelfth international
  conference on {Information} and knowledge management}}
  \emph{(\bibinfo{series}{{CIKM} '03})}. \bibinfo{publisher}{Association for
  Computing Machinery}, \bibinfo{address}{New York, NY, USA},
  \bibinfo{pages}{556--559}.
\newblock
\showISBNx{978-1-58113-723-1}
\urldef\tempurl%
\url{https://doi.org/10.1145/956863.956972}
\showDOI{\tempurl}


\bibitem[Lu and Zhou(2011)]%
        {lu_link_2011}
\bibfield{author}{\bibinfo{person}{Linyuan Lu} {and} \bibinfo{person}{Tao
  Zhou}.} \bibinfo{year}{2011}\natexlab{}.
\newblock \showarticletitle{Link {Prediction} in {Complex} {Networks}: {A}
  {Survey}}.
\newblock \bibinfo{journal}{\emph{Physica A: Statistical Mechanics and its
  Applications}} \bibinfo{volume}{390}, \bibinfo{number}{6}
  (\bibinfo{date}{March} \bibinfo{year}{2011}), \bibinfo{pages}{1150--1170}.
\newblock
\showISSN{03784371}
\urldef\tempurl%
\url{https://doi.org/10.1016/j.physa.2010.11.027}
\showDOI{\tempurl}
\newblock
\shownote{arXiv:1010.0725 [physics]}.


\bibitem[Luo et~al\mbox{.}(2023)]%
        {luo_automated_2023}
\bibfield{author}{\bibinfo{person}{Youzhi Luo}, \bibinfo{person}{Michael
  McThrow}, \bibinfo{person}{Wing~Yee Au}, \bibinfo{person}{Tao Komikado},
  \bibinfo{person}{Kanji Uchino}, \bibinfo{person}{Koji Maruhashi}, {and}
  \bibinfo{person}{Shuiwang Ji}.} \bibinfo{year}{2023}\natexlab{}.
\newblock \bibinfo{title}{Automated {Data} {Augmentations} for {Graph}
  {Classification}}.
\newblock
\newblock
\urldef\tempurl%
\url{https://doi.org/10.48550/arXiv.2202.13248}
\showDOI{\tempurl}
\newblock
\shownote{arXiv:2202.13248 [cs]}.


\bibitem[Maddison et~al\mbox{.}(2023)]%
        {maddison_concrete_2023}
\bibfield{author}{\bibinfo{person}{Chris~J. Maddison}, \bibinfo{person}{Andriy
  Mnih}, {and} \bibinfo{person}{Yee~Whye Teh}.}
  \bibinfo{year}{2023}\natexlab{}.
\newblock \showarticletitle{The {Concrete} {Distribution}: {A} {Continuous}
  {Relaxation} of {Discrete} {Random} {Variables}}.
\newblock
\urldef\tempurl%
\url{https://openreview.net/forum?id=S1jE5L5gl}
\showURL{%
\tempurl}


\bibitem[Miao et~al\mbox{.}(2022)]%
        {miao_interpretable_2022}
\bibfield{author}{\bibinfo{person}{Siqi Miao}, \bibinfo{person}{Miaoyuan Liu},
  {and} \bibinfo{person}{Pan Li}.} \bibinfo{year}{2022}\natexlab{}.
\newblock \bibinfo{title}{Interpretable and {Generalizable} {Graph} {Learning}
  via {Stochastic} {Attention} {Mechanism}}.
\newblock
\newblock
\urldef\tempurl%
\url{https://doi.org/10.48550/arXiv.2201.12987}
\showDOI{\tempurl}
\newblock
\shownote{arXiv:2201.12987 [cs]}.


\bibitem[Newman(2006)]%
        {newman_finding_2006}
\bibfield{author}{\bibinfo{person}{Mark~EJ Newman}.}
  \bibinfo{year}{2006}\natexlab{}.
\newblock \showarticletitle{Finding community structure in networks using the
  eigenvectors of matrices}.
\newblock \bibinfo{journal}{\emph{Physical review E}} \bibinfo{volume}{74},
  \bibinfo{number}{3} (\bibinfo{year}{2006}), \bibinfo{pages}{036104}.
\newblock
\newblock
\shownote{Publisher: APS}.


\bibitem[Pan et~al\mbox{.}(2022)]%
        {pan_neural_2022}
\bibfield{author}{\bibinfo{person}{Liming Pan}, \bibinfo{person}{Cheng Shi},
  {and} \bibinfo{person}{Ivan Dokmanić}.} \bibinfo{year}{2022}\natexlab{}.
\newblock \showarticletitle{Neural {Link} {Prediction} with {Walk} {Pooling}}.
  In \bibinfo{booktitle}{\emph{International {Conference} on {Learning}
  {Representations}}}.
\newblock
\urldef\tempurl%
\url{https://openreview.net/forum?id=CCu6RcUMwK0}
\showURL{%
\tempurl}


\bibitem[Papp et~al\mbox{.}(2021)]%
        {papp_dropgnn_2021}
\bibfield{author}{\bibinfo{person}{Pál~András Papp}, \bibinfo{person}{Karolis
  Martinkus}, \bibinfo{person}{Lukas Faber}, {and} \bibinfo{person}{Roger
  Wattenhofer}.} \bibinfo{year}{2021}\natexlab{}.
\newblock \bibinfo{title}{{DropGNN}: {Random} {Dropouts} {Increase} the
  {Expressiveness} of {Graph} {Neural} {Networks}}.
\newblock
\newblock
\urldef\tempurl%
\url{http://arxiv.org/abs/2111.06283}
\showURL{%
\tempurl}
\newblock
\shownote{arXiv:2111.06283 [cs]}.


\bibitem[Rong et~al\mbox{.}(2020)]%
        {rong_dropedge_2020}
\bibfield{author}{\bibinfo{person}{Yu Rong}, \bibinfo{person}{Wenbing Huang},
  \bibinfo{person}{Tingyang Xu}, {and} \bibinfo{person}{Junzhou Huang}.}
  \bibinfo{year}{2020}\natexlab{}.
\newblock \showarticletitle{{DropEdge}: {Towards} {Deep} {Graph}
  {Convolutional} {Networks} on {Node} {Classification}}. In
  \bibinfo{booktitle}{\emph{International {Conference} on {Learning}
  {Representations}}}.
\newblock
\urldef\tempurl%
\url{https://openreview.net/forum?id=Hkx1qkrKPr}
\showURL{%
\tempurl}


\bibitem[Shchur et~al\mbox{.}(2019)]%
        {shchur_pitfalls_2019}
\bibfield{author}{\bibinfo{person}{Oleksandr Shchur},
  \bibinfo{person}{Maximilian Mumme}, \bibinfo{person}{Aleksandar Bojchevski},
  {and} \bibinfo{person}{Stephan Günnemann}.} \bibinfo{year}{2019}\natexlab{}.
\newblock \bibinfo{title}{Pitfalls of {Graph} {Neural} {Network} {Evaluation}}.
\newblock
\newblock
\urldef\tempurl%
\url{https://doi.org/10.48550/arXiv.1811.05868}
\showDOI{\tempurl}
\newblock
\shownote{arXiv:1811.05868 [cs, stat]}.


\bibitem[Shorten and Khoshgoftaar(2019)]%
        {shorten_survey_2019}
\bibfield{author}{\bibinfo{person}{Connor Shorten} {and}
  \bibinfo{person}{Taghi~M. Khoshgoftaar}.} \bibinfo{year}{2019}\natexlab{}.
\newblock \showarticletitle{A survey on {Image} {Data} {Augmentation} for
  {Deep} {Learning}}.
\newblock \bibinfo{journal}{\emph{Journal of Big Data}} \bibinfo{volume}{6},
  \bibinfo{number}{1} (\bibinfo{date}{July} \bibinfo{year}{2019}),
  \bibinfo{pages}{60}.
\newblock
\showISSN{2196-1115}
\urldef\tempurl%
\url{https://doi.org/10.1186/s40537-019-0197-0}
\showDOI{\tempurl}


\bibitem[Singh et~al\mbox{.}(2021)]%
        {singh_edge_2021}
\bibfield{author}{\bibinfo{person}{Abhay Singh}, \bibinfo{person}{Qian Huang},
  \bibinfo{person}{Sijia~Linda Huang}, \bibinfo{person}{Omkar Bhalerao},
  \bibinfo{person}{Horace He}, \bibinfo{person}{Ser-Nam Lim}, {and}
  \bibinfo{person}{Austin~R. Benson}.} \bibinfo{year}{2021}\natexlab{}.
\newblock \bibinfo{title}{Edge {Proposal} {Sets} for {Link} {Prediction}}.
\newblock
\newblock
\urldef\tempurl%
\url{https://doi.org/10.48550/arXiv.2106.15810}
\showDOI{\tempurl}
\newblock
\shownote{arXiv:2106.15810 [cs]}.


\bibitem[Singh et~al\mbox{.}(2018)]%
        {singh_hide-and-seek_2018}
\bibfield{author}{\bibinfo{person}{Krishna~Kumar Singh}, \bibinfo{person}{Hao
  Yu}, \bibinfo{person}{Aron Sarmasi}, \bibinfo{person}{Gautam Pradeep}, {and}
  \bibinfo{person}{Yong~Jae Lee}.} \bibinfo{year}{2018}\natexlab{}.
\newblock \bibinfo{title}{Hide-and-{Seek}: {A} {Data} {Augmentation}
  {Technique} for {Weakly}-{Supervised} {Localization} and {Beyond}}.
\newblock
\newblock
\urldef\tempurl%
\url{https://doi.org/10.48550/arXiv.1811.02545}
\showDOI{\tempurl}
\newblock
\shownote{arXiv:1811.02545 [cs]}.


\bibitem[Spring et~al\mbox{.}(2002)]%
        {spring_measuring_2002}
\bibfield{author}{\bibinfo{person}{Neil Spring}, \bibinfo{person}{Ratul
  Mahajan}, {and} \bibinfo{person}{David Wetherall}.}
  \bibinfo{year}{2002}\natexlab{}.
\newblock \showarticletitle{Measuring {ISP} topologies with {Rocketfuel}}.
\newblock \bibinfo{journal}{\emph{ACM SIGCOMM Computer Communication Review}}
  \bibinfo{volume}{32}, \bibinfo{number}{4} (\bibinfo{year}{2002}),
  \bibinfo{pages}{133--145}.
\newblock
\newblock
\shownote{Publisher: ACM New York, NY, USA}.


\bibitem[Sun et~al\mbox{.}(2021)]%
        {sun_graph_2021}
\bibfield{author}{\bibinfo{person}{Qingyun Sun}, \bibinfo{person}{Jianxin Li},
  \bibinfo{person}{Hao Peng}, \bibinfo{person}{Jia Wu},
  \bibinfo{person}{Xingcheng Fu}, \bibinfo{person}{Cheng Ji}, {and}
  \bibinfo{person}{Philip~S. Yu}.} \bibinfo{year}{2021}\natexlab{}.
\newblock \bibinfo{title}{Graph {Structure} {Learning} with {Variational}
  {Information} {Bottleneck}}.
\newblock
\newblock
\urldef\tempurl%
\url{http://arxiv.org/abs/2112.08903}
\showURL{%
\tempurl}
\newblock
\shownote{arXiv:2112.08903 [cs]}.


\bibitem[Suresh et~al\mbox{.}(2021)]%
        {suresh_adversarial_2021}
\bibfield{author}{\bibinfo{person}{Susheel Suresh}, \bibinfo{person}{Pan Li},
  \bibinfo{person}{Cong Hao}, {and} \bibinfo{person}{Jennifer Neville}.}
  \bibinfo{year}{2021}\natexlab{}.
\newblock \bibinfo{title}{Adversarial {Graph} {Augmentation} to {Improve}
  {Graph} {Contrastive} {Learning}}.
\newblock
\newblock
\urldef\tempurl%
\url{https://doi.org/10.48550/arXiv.2106.05819}
\showDOI{\tempurl}
\newblock
\shownote{arXiv:2106.05819 [cs]}.


\bibitem[Szklarczyk et~al\mbox{.}(2019)]%
        {szklarczyk_string_2019}
\bibfield{author}{\bibinfo{person}{Damian Szklarczyk},
  \bibinfo{person}{Annika~L. Gable}, \bibinfo{person}{David Lyon},
  \bibinfo{person}{Alexander Junge}, \bibinfo{person}{Stefan Wyder},
  \bibinfo{person}{Jaime Huerta-Cepas}, \bibinfo{person}{Milan Simonovic},
  \bibinfo{person}{Nadezhda~T. Doncheva}, \bibinfo{person}{John~H. Morris},
  \bibinfo{person}{Peer Bork}, \bibinfo{person}{Lars~J. Jensen}, {and}
  \bibinfo{person}{Christian~von Mering}.} \bibinfo{year}{2019}\natexlab{}.
\newblock \showarticletitle{{STRING} v11: protein-protein association networks
  with increased coverage, supporting functional discovery in genome-wide
  experimental datasets}.
\newblock \bibinfo{journal}{\emph{Nucleic Acids Research}}
  \bibinfo{volume}{47}, \bibinfo{number}{D1} (\bibinfo{date}{Jan.}
  \bibinfo{year}{2019}), \bibinfo{pages}{D607--D613}.
\newblock
\showISSN{1362-4962}
\urldef\tempurl%
\url{https://doi.org/10.1093/nar/gky1131}
\showDOI{\tempurl}


\bibitem[Tishby et~al\mbox{.}(2000)]%
        {tishby_information_2000}
\bibfield{author}{\bibinfo{person}{Naftali Tishby},
  \bibinfo{person}{Fernando~C. Pereira}, {and} \bibinfo{person}{William
  Bialek}.} \bibinfo{year}{2000}\natexlab{}.
\newblock \bibinfo{title}{The information bottleneck method}.
\newblock
\newblock
\urldef\tempurl%
\url{https://doi.org/10.48550/arXiv.physics/0004057}
\showDOI{\tempurl}
\newblock
\shownote{arXiv:physics/0004057}.


\bibitem[Tishby and Zaslavsky(2015)]%
        {tishby_deep_2015}
\bibfield{author}{\bibinfo{person}{Naftali Tishby} {and} \bibinfo{person}{Noga
  Zaslavsky}.} \bibinfo{year}{2015}\natexlab{}.
\newblock \showarticletitle{Deep learning and the information bottleneck
  principle}. In \bibinfo{booktitle}{\emph{2015 {IEEE} {Information} {Theory}
  {Workshop} ({ITW})}}. \bibinfo{pages}{1--5}.
\newblock
\urldef\tempurl%
\url{https://doi.org/10.1109/ITW.2015.7133169}
\showDOI{\tempurl}


\bibitem[Topping et~al\mbox{.}(2022)]%
        {topping_understanding_2022}
\bibfield{author}{\bibinfo{person}{Jake Topping}, \bibinfo{person}{Francesco
  Di~Giovanni}, \bibinfo{person}{Benjamin~Paul Chamberlain},
  \bibinfo{person}{Xiaowen Dong}, {and} \bibinfo{person}{Michael~M.
  Bronstein}.} \bibinfo{year}{2022}\natexlab{}.
\newblock \showarticletitle{Understanding over-squashing and bottlenecks on
  graphs via curvature}.
\newblock \bibinfo{journal}{\emph{arXiv:2111.14522 [cs, stat]}}
  (\bibinfo{date}{March} \bibinfo{year}{2022}).
\newblock
\urldef\tempurl%
\url{http://arxiv.org/abs/2111.14522}
\showURL{%
\tempurl}
\newblock
\shownote{arXiv: 2111.14522}.


\bibitem[Vaswani et~al\mbox{.}(2017)]%
        {vaswani_attention_2017}
\bibfield{author}{\bibinfo{person}{Ashish Vaswani}, \bibinfo{person}{Noam
  Shazeer}, \bibinfo{person}{Niki Parmar}, \bibinfo{person}{Jakob Uszkoreit},
  \bibinfo{person}{Llion Jones}, \bibinfo{person}{Aidan~N Gomez},
  \bibinfo{person}{Lukasz Kaiser}, {and} \bibinfo{person}{Illia Polosukhin}.}
  \bibinfo{year}{2017}\natexlab{}.
\newblock \showarticletitle{Attention is {All} you {Need}}. In
  \bibinfo{booktitle}{\emph{Advances in {Neural} {Information} {Processing}
  {Systems}}}, Vol.~\bibinfo{volume}{30}. \bibinfo{publisher}{Curran
  Associates, Inc.}
\newblock
\urldef\tempurl%
\url{https://papers.nips.cc/paper/2017/hash/3f5ee243547dee91fbd053c1c4a845aa-Abstract.html}
\showURL{%
\tempurl}


\bibitem[Veličković et~al\mbox{.}(2018)]%
        {velickovic_graph_2018}
\bibfield{author}{\bibinfo{person}{Petar Veličković},
  \bibinfo{person}{Guillem Cucurull}, \bibinfo{person}{Arantxa Casanova},
  \bibinfo{person}{Adriana Romero}, \bibinfo{person}{Pietro Liò}, {and}
  \bibinfo{person}{Yoshua Bengio}.} \bibinfo{year}{2018}\natexlab{}.
\newblock \showarticletitle{Graph {Attention} {Networks}}.
\newblock \bibinfo{journal}{\emph{arXiv:1710.10903 [cs, stat]}}
  (\bibinfo{date}{Feb.} \bibinfo{year}{2018}).
\newblock
\urldef\tempurl%
\url{http://arxiv.org/abs/1710.10903}
\showURL{%
\tempurl}
\newblock
\shownote{arXiv: 1710.10903}.


\bibitem[Von~Mering et~al\mbox{.}(2002)]%
        {von_mering_comparative_2002}
\bibfield{author}{\bibinfo{person}{Christian Von~Mering},
  \bibinfo{person}{Roland Krause}, \bibinfo{person}{Berend Snel},
  \bibinfo{person}{Michael Cornell}, \bibinfo{person}{Stephen~G Oliver},
  \bibinfo{person}{Stanley Fields}, {and} \bibinfo{person}{Peer Bork}.}
  \bibinfo{year}{2002}\natexlab{}.
\newblock \showarticletitle{Comparative assessment of large-scale data sets of
  protein–protein interactions}.
\newblock \bibinfo{journal}{\emph{Nature}} \bibinfo{volume}{417},
  \bibinfo{number}{6887} (\bibinfo{year}{2002}), \bibinfo{pages}{399--403}.
\newblock
\newblock
\shownote{Publisher: Nature Publishing Group}.


\bibitem[Wang et~al\mbox{.}(2023)]%
        {wang_neural_2023}
\bibfield{author}{\bibinfo{person}{Xiyuan Wang}, \bibinfo{person}{Haotong
  Yang}, {and} \bibinfo{person}{Muhan Zhang}.} \bibinfo{year}{2023}\natexlab{}.
\newblock \bibinfo{title}{Neural {Common} {Neighbor} with {Completion} for
  {Link} {Prediction}}.
\newblock
\newblock
\urldef\tempurl%
\url{https://doi.org/10.48550/arXiv.2302.00890}
\showDOI{\tempurl}
\newblock
\shownote{arXiv:2302.00890 [cs]}.


\bibitem[Wang et~al\mbox{.}(2020)]%
        {wang_nodeaug_2020}
\bibfield{author}{\bibinfo{person}{Yiwei Wang}, \bibinfo{person}{Wei Wang},
  \bibinfo{person}{Yuxuan Liang}, \bibinfo{person}{Yujun Cai},
  \bibinfo{person}{Juncheng Liu}, {and} \bibinfo{person}{Bryan Hooi}.}
  \bibinfo{year}{2020}\natexlab{}.
\newblock \showarticletitle{{NodeAug}: {Semi}-{Supervised} {Node}
  {Classification} with {Data} {Augmentation}}. In
  \bibinfo{booktitle}{\emph{Proceedings of the 26th {ACM} {SIGKDD}
  {International} {Conference} on {Knowledge} {Discovery} \& {Data} {Mining}}}
  \emph{(\bibinfo{series}{{KDD} '20})}. \bibinfo{publisher}{Association for
  Computing Machinery}, \bibinfo{address}{New York, NY, USA},
  \bibinfo{pages}{207--217}.
\newblock
\showISBNx{978-1-4503-7998-4}
\urldef\tempurl%
\url{https://doi.org/10.1145/3394486.3403063}
\showDOI{\tempurl}


\bibitem[Watts and Strogatz(1998)]%
        {watts_collective_1998}
\bibfield{author}{\bibinfo{person}{Duncan~J. Watts} {and}
  \bibinfo{person}{Steven~H. Strogatz}.} \bibinfo{year}{1998}\natexlab{}.
\newblock \showarticletitle{Collective dynamics of ‘small-world’ networks}.
\newblock \bibinfo{journal}{\emph{Nature}}  \bibinfo{volume}{393}
  (\bibinfo{year}{1998}), \bibinfo{pages}{440--442}.
\newblock
\urldef\tempurl%
\url{https://api.semanticscholar.org/CorpusID:3034643}
\showURL{%
\tempurl}


\bibitem[Wu et~al\mbox{.}(2022)]%
        {wu_handling_2022}
\bibfield{author}{\bibinfo{person}{Qitian Wu}, \bibinfo{person}{Hengrui Zhang},
  \bibinfo{person}{Junchi Yan}, {and} \bibinfo{person}{David Wipf}.}
  \bibinfo{year}{2022}\natexlab{}.
\newblock \showarticletitle{Handling {Distribution} {Shifts} on {Graphs}: {An}
  {Invariance} {Perspective}}.
\newblock
\urldef\tempurl%
\url{https://openreview.net/forum?id=FQOC5u-1egI}
\showURL{%
\tempurl}


\bibitem[Wu et~al\mbox{.}(2020)]%
        {wu_graph_2020}
\bibfield{author}{\bibinfo{person}{Tailin Wu}, \bibinfo{person}{Hongyu Ren},
  \bibinfo{person}{Pan Li}, {and} \bibinfo{person}{Jure Leskovec}.}
  \bibinfo{year}{2020}\natexlab{}.
\newblock \bibinfo{title}{Graph {Information} {Bottleneck}}.
\newblock
\newblock
\urldef\tempurl%
\url{http://arxiv.org/abs/2010.12811}
\showURL{%
\tempurl}
\newblock
\shownote{arXiv:2010.12811 [cs, stat]}.


\bibitem[Xie et~al\mbox{.}(2020)]%
        {xie_unsupervised_2020}
\bibfield{author}{\bibinfo{person}{Qizhe Xie}, \bibinfo{person}{Zihang Dai},
  \bibinfo{person}{Eduard Hovy}, \bibinfo{person}{Minh-Thang Luong}, {and}
  \bibinfo{person}{Quoc~V. Le}.} \bibinfo{year}{2020}\natexlab{}.
\newblock \bibinfo{title}{Unsupervised {Data} {Augmentation} for {Consistency}
  {Training}}.
\newblock
\newblock
\urldef\tempurl%
\url{https://doi.org/10.48550/arXiv.1904.12848}
\showDOI{\tempurl}
\newblock
\shownote{arXiv:1904.12848 [cs, stat]}.


\bibitem[Xu et~al\mbox{.}(2018)]%
        {xu_how_2018}
\bibfield{author}{\bibinfo{person}{Keyulu Xu}, \bibinfo{person}{Weihua Hu},
  \bibinfo{person}{Jure Leskovec}, {and} \bibinfo{person}{Stefanie Jegelka}.}
  \bibinfo{year}{2018}\natexlab{}.
\newblock \showarticletitle{How {Powerful} are {Graph} {Neural} {Networks}?}
\newblock \bibinfo{journal}{\emph{CoRR}}  \bibinfo{volume}{abs/1810.00826}
  (\bibinfo{year}{2018}).
\newblock
\urldef\tempurl%
\url{http://arxiv.org/abs/1810.00826}
\showURL{%
\tempurl}
\newblock
\shownote{arXiv: 1810.00826}.


\bibitem[You et~al\mbox{.}(2021)]%
        {you_graph_2021}
\bibfield{author}{\bibinfo{person}{Yuning You}, \bibinfo{person}{Tianlong
  Chen}, \bibinfo{person}{Yongduo Sui}, \bibinfo{person}{Ting Chen},
  \bibinfo{person}{Zhangyang Wang}, {and} \bibinfo{person}{Yang Shen}.}
  \bibinfo{year}{2021}\natexlab{}.
\newblock \bibinfo{title}{Graph {Contrastive} {Learning} with {Augmentations}}.
\newblock
\newblock
\urldef\tempurl%
\url{https://doi.org/10.48550/arXiv.2010.13902}
\showDOI{\tempurl}
\newblock
\shownote{arXiv:2010.13902 [cs]}.


\bibitem[Yu et~al\mbox{.}(2020)]%
        {yu_graph_2020}
\bibfield{author}{\bibinfo{person}{Junchi Yu}, \bibinfo{person}{Tingyang Xu},
  \bibinfo{person}{Yu Rong}, \bibinfo{person}{Yatao Bian},
  \bibinfo{person}{Junzhou Huang}, {and} \bibinfo{person}{Ran He}.}
  \bibinfo{year}{2020}\natexlab{}.
\newblock \bibinfo{title}{Graph {Information} {Bottleneck} for {Subgraph}
  {Recognition}}.
\newblock
\newblock
\urldef\tempurl%
\url{https://doi.org/10.48550/arXiv.2010.05563}
\showDOI{\tempurl}
\newblock
\shownote{arXiv:2010.05563 [cs, stat]}.


\bibitem[Zhang and Chen(2018)]%
        {zhang_link_2018}
\bibfield{author}{\bibinfo{person}{Muhan Zhang} {and} \bibinfo{person}{Yixin
  Chen}.} \bibinfo{year}{2018}\natexlab{}.
\newblock \showarticletitle{Link {Prediction} {Based} on {Graph} {Neural}
  {Networks}}. In \bibinfo{booktitle}{\emph{Advances in {Neural} {Information}
  {Processing} {Systems}}}, \bibfield{editor}{\bibinfo{person}{S.~Bengio},
  \bibinfo{person}{H.~Wallach}, \bibinfo{person}{H.~Larochelle},
  \bibinfo{person}{K.~Grauman}, \bibinfo{person}{N.~Cesa-Bianchi}, {and}
  \bibinfo{person}{R.~Garnett}} (Eds.), Vol.~\bibinfo{volume}{31}.
  \bibinfo{publisher}{Curran Associates, Inc.}
\newblock
\urldef\tempurl%
\url{https://proceedings.neurips.cc/paper/2018/file/53f0d7c537d99b3824f0f99d62ea2428-Paper.pdf}
\showURL{%
\tempurl}


\bibitem[Zhang et~al\mbox{.}(2021)]%
        {zhang_labeling_2021}
\bibfield{author}{\bibinfo{person}{Muhan Zhang}, \bibinfo{person}{Pan Li},
  \bibinfo{person}{Yinglong Xia}, \bibinfo{person}{Kai Wang}, {and}
  \bibinfo{person}{Long Jin}.} \bibinfo{year}{2021}\natexlab{}.
\newblock \showarticletitle{Labeling {Trick}: {A} {Theory} of {Using} {Graph}
  {Neural} {Networks} for {Multi}-{Node} {Representation} {Learning}}. In
  \bibinfo{booktitle}{\emph{Advances in {Neural} {Information} {Processing}
  {Systems}}}, \bibfield{editor}{\bibinfo{person}{M.~Ranzato},
  \bibinfo{person}{A.~Beygelzimer}, \bibinfo{person}{Y.~Dauphin},
  \bibinfo{person}{P.~S. Liang}, {and} \bibinfo{person}{J.~Wortman Vaughan}}
  (Eds.), Vol.~\bibinfo{volume}{34}. \bibinfo{publisher}{Curran Associates,
  Inc.}, \bibinfo{pages}{9061--9073}.
\newblock
\urldef\tempurl%
\url{https://proceedings.neurips.cc/paper/2021/file/4be49c79f233b4f4070794825c323733-Paper.pdf}
\showURL{%
\tempurl}


\bibitem[Zhang et~al\mbox{.}(2022)]%
        {zhang_unsupervised_2022}
\bibfield{author}{\bibinfo{person}{Sixiao Zhang}, \bibinfo{person}{Hongxu
  Chen}, \bibinfo{person}{Xiangguo Sun}, \bibinfo{person}{Yicong Li}, {and}
  \bibinfo{person}{Guandong Xu}.} \bibinfo{year}{2022}\natexlab{}.
\newblock \showarticletitle{Unsupervised {Graph} {Poisoning} {Attack} via
  {Contrastive} {Loss} {Back}-propagation}. In
  \bibinfo{booktitle}{\emph{Proceedings of the {ACM} {Web} {Conference} 2022}}.
  \bibinfo{pages}{1322--1330}.
\newblock
\urldef\tempurl%
\url{https://doi.org/10.1145/3485447.3512179}
\showDOI{\tempurl}
\newblock
\shownote{arXiv:2201.07986 [cs]}.


\bibitem[Zhang et~al\mbox{.}(2016)]%
        {zhang_character-level_2016}
\bibfield{author}{\bibinfo{person}{Xiang Zhang}, \bibinfo{person}{Junbo Zhao},
  {and} \bibinfo{person}{Yann LeCun}.} \bibinfo{year}{2016}\natexlab{}.
\newblock \bibinfo{title}{Character-level {Convolutional} {Networks} for {Text}
  {Classification}}.
\newblock
\newblock
\urldef\tempurl%
\url{https://doi.org/10.48550/arXiv.1509.01626}
\showDOI{\tempurl}
\newblock
\shownote{arXiv:1509.01626 [cs]}.


\bibitem[Zhao et~al\mbox{.}(2023)]%
        {zhao_graph_2023}
\bibfield{author}{\bibinfo{person}{Tong Zhao}, \bibinfo{person}{Wei Jin},
  \bibinfo{person}{Yozen Liu}, \bibinfo{person}{Yingheng Wang},
  \bibinfo{person}{Gang Liu}, \bibinfo{person}{Stephan Günnemann},
  \bibinfo{person}{Neil Shah}, {and} \bibinfo{person}{Meng Jiang}.}
  \bibinfo{year}{2023}\natexlab{}.
\newblock \bibinfo{title}{Graph {Data} {Augmentation} for {Graph} {Machine}
  {Learning}: {A} {Survey}}.
\newblock
\newblock
\urldef\tempurl%
\url{https://doi.org/10.48550/arXiv.2202.08871}
\showDOI{\tempurl}
\newblock
\shownote{arXiv:2202.08871 [cs]}.


\bibitem[Zhao et~al\mbox{.}(2022)]%
        {zhao_learning_2022}
\bibfield{author}{\bibinfo{person}{Tong Zhao}, \bibinfo{person}{Gang Liu},
  \bibinfo{person}{Daheng Wang}, \bibinfo{person}{Wenhao Yu}, {and}
  \bibinfo{person}{Meng Jiang}.} \bibinfo{year}{2022}\natexlab{}.
\newblock \showarticletitle{Learning from {Counterfactual} {Links} for {Link}
  {Prediction}}. In \bibinfo{booktitle}{\emph{Proceedings of the 39th
  {International} {Conference} on {Machine} {Learning}}}
  \emph{(\bibinfo{series}{Proceedings of {Machine} {Learning} {Research}},
  Vol.~\bibinfo{volume}{162})}, \bibfield{editor}{\bibinfo{person}{Kamalika
  Chaudhuri}, \bibinfo{person}{Stefanie Jegelka}, \bibinfo{person}{Le~Song},
  \bibinfo{person}{Csaba Szepesvari}, \bibinfo{person}{Gang Niu}, {and}
  \bibinfo{person}{Sivan Sabato}} (Eds.). \bibinfo{publisher}{PMLR},
  \bibinfo{pages}{26911--26926}.
\newblock
\urldef\tempurl%
\url{https://proceedings.mlr.press/v162/zhao22e.html}
\showURL{%
\tempurl}


\bibitem[Zheng et~al\mbox{.}(2020)]%
        {zheng_robust_2020}
\bibfield{author}{\bibinfo{person}{Cheng Zheng}, \bibinfo{person}{Bo Zong},
  \bibinfo{person}{Wei Cheng}, \bibinfo{person}{Dongjin Song},
  \bibinfo{person}{Jingchao Ni}, \bibinfo{person}{Wenchao Yu},
  \bibinfo{person}{Haifeng Chen}, {and} \bibinfo{person}{Wei Wang}.}
  \bibinfo{year}{2020}\natexlab{}.
\newblock \showarticletitle{Robust {Graph} {Representation} {Learning} via
  {Neural} {Sparsification}}. In \bibinfo{booktitle}{\emph{Proceedings of the
  37th {International} {Conference} on {Machine} {Learning}}}.
  \bibinfo{publisher}{PMLR}, \bibinfo{pages}{11458--11468}.
\newblock
\urldef\tempurl%
\url{https://proceedings.mlr.press/v119/zheng20d.html}
\showURL{%
\tempurl}
\newblock
\shownote{ISSN: 2640-3498}.


\bibitem[Zhong et~al\mbox{.}(2017)]%
        {zhong_random_2017}
\bibfield{author}{\bibinfo{person}{Zhun Zhong}, \bibinfo{person}{Liang Zheng},
  \bibinfo{person}{Guoliang Kang}, \bibinfo{person}{Shaozi Li}, {and}
  \bibinfo{person}{Yi Yang}.} \bibinfo{year}{2017}\natexlab{}.
\newblock \bibinfo{title}{Random {Erasing} {Data} {Augmentation}}.
\newblock
\newblock
\urldef\tempurl%
\url{https://doi.org/10.48550/arXiv.1708.04896}
\showDOI{\tempurl}
\newblock
\shownote{arXiv:1708.04896 [cs]}.


\bibitem[Zhou et~al\mbox{.}(2009)]%
        {zhou_predicting_2009}
\bibfield{author}{\bibinfo{person}{Tao Zhou}, \bibinfo{person}{Linyuan Lü},
  {and} \bibinfo{person}{Yi-Cheng Zhang}.} \bibinfo{year}{2009}\natexlab{}.
\newblock \showarticletitle{Predicting missing links via local information}.
\newblock \bibinfo{journal}{\emph{The European Physical Journal B}}
  \bibinfo{volume}{71}, \bibinfo{number}{4} (\bibinfo{year}{2009}),
  \bibinfo{pages}{623--630}.
\newblock
\newblock
\shownote{Publisher: Springer}.


\bibitem[Zügner et~al\mbox{.}(2018)]%
        {zugner_adversarial_2018}
\bibfield{author}{\bibinfo{person}{Daniel Zügner}, \bibinfo{person}{Amir
  Akbarnejad}, {and} \bibinfo{person}{Stephan Günnemann}.}
  \bibinfo{year}{2018}\natexlab{}.
\newblock \showarticletitle{Adversarial {Attacks} on {Neural} {Networks} for
  {Graph} {Data}}. In \bibinfo{booktitle}{\emph{Proceedings of the 24th {ACM}
  {SIGKDD} {International} {Conference} on {Knowledge} {Discovery} \&amp;
  {Data} {Mining}}} \emph{(\bibinfo{series}{{KDD} '18})}.
  \bibinfo{publisher}{Association for Computing Machinery},
  \bibinfo{address}{New York, NY, USA}, \bibinfo{pages}{2847--2856}.
\newblock
\showISBNx{978-1-4503-5552-0}
\urldef\tempurl%
\url{https://doi.org/10.1145/3219819.3220078}
\showDOI{\tempurl}
\newblock
\shownote{event-place: London, United Kingdom}.


\end{thebibliography}

\clearpage
\appendix
\section{Related works}
\paragraph{Data augmentation.} 
The efficacy of data augmentation strategies in enhancing model generalization is well-documented across various domains. 
Traditional DA techniques, such as oversampling, undersampling, and interpolation methods~\cite{chawla_smote_2002}, 
have proven to be instrumental in mitigating issues related to learning from imbalanced datasets. 
In recent years, DA has found extensive application in computer vision and natural language processing. 
Within the realm of computer vision, techniques such as horizontal flipping, random erasing~\cite{zhong_random_2017}, Hide-and-Seek~\cite{singh_hide-and-seek_2018}, and Cutout~\cite{devries_improved_2017} have demonstrated their ability to bolster model performance. 
On the other hand, in natural language processing, DA is often achieved through lexical substitution strategies, where words are replaced with their semantically equivalent counterparts~\cite{zhang_character-level_2016}. 
UDA~\cite{xie_unsupervised_2020} introduces a novel approach that 
leverages TF-IDF metrics for keyword feature augmentation in documents.

\paragraph{Graph data augmentation.} 
Graph-structured data, with its heterogeneous information modalities and complex properties, 
presents a more intricate landscape for DA compared to conventional image or text data. 
Typically, graph data augmentation can be bifurcated into two primary approaches: perturbation of graph structure and enhancement of node attributes.

In the realm of semi-supervised node classification, several innovative techniques have been proposed. Drop Edge~\cite{rong_dropedge_2020}, for instance, introduces random edge dropping to mitigate the oversmoothing problem prevalent in GNNs. 
Similarly, SDRF~\cite{topping_understanding_2022} leverages graph structure rewiring to address the over-squashing issue in GNNs. 
NodeAug~\cite{wang_nodeaug_2020} proposes a more holistic approach by simultaneously augmenting both the graph structure (via edge addition/deletion) and node attributes (via feature replacement).

As for the graph classification task, AD-GCL~\cite{suresh_adversarial_2021} pioneers an adversarial augmentation technique to boost the training of graph contrastive learning. 
Concurrently, JOAO~\cite{you_graph_2021} and GraphAug~\cite{luo_automated_2023} automates the selection of augmentations from a predefined pool, incorporating both edge perturbation and node attribute masking.

However, the domain of link prediction has seen relatively limited exploration of DA. Notable exceptions include Distance Encoding~\cite{li_distance_2020} and Node labeling~\cite{zhang_labeling_2021}, which enhance GNNs' expressiveness by infusing distance information. 
Moreover, \citet{hwang_analysis_2022} proposes to improve both model expressiveness and node impact by incorporating a virtual node 
as a message-passing hub for link prediction.

\paragraph{Information bottleneck principle.} 
The Information Bottleneck (IB) principle has been increasingly incorporated into deep learning models to enhance learning robustness. DeepVIB~\cite{alemi_deep_2023}, for instance, fistly introduces the application of the IB principle in this domain. 
To overcome the intractable computation posed by the mutual information term in IB, 
DeepVIB devises a variational approximation akin to the approach used in Variational Autoencoders (VAEs)~\cite{kingma_auto-encoding_2014}.
The principle of IB has also been leveraged within the realm of graph representation learning. 
GIB~\cite{wu_graph_2020} was among the first to integrate IB into graph learning, aiming to protect GNNs from adversarial attacks. 
VIB-GSL~\cite{sun_graph_2021} expanded upon this by applying the IB principle to graph structure learning, demonstrating its robustness in graph classification tasks.
GSAT~\cite{miao_interpretable_2022}, which is the most related to our work, employs IB to extract the most rationale components from graphs for interpretation purposes. 
Similarly, IB-subgraph~\cite{yu_graph_2020} uses IB in conjunction with a bi-level optimization process to identify the most representative subgraph components.

\subsection{Comparison to GSAT~\cite{miao_interpretable_2022}}
\paragraph{Scope \& Objective}
While GSAT has significantly influenced our research, our study introduces a novel graph DA method designed specifically for link prediction tasks, distinguishing itself from GSAT's focus on graph classification interpretability.

\paragraph{Strategic Design}
Unlike the direct application of GIB/GSAT to link prediction, we strategically designed \our's components:

\begin{itemize}
    \item \textbf{Isolated DA Design}: GSAT when directly applied to the link prediction task resulted in conflicts between optimal DAs for varying target links. We address this by adopting a subgraph-based approach for isolating DA effects per target link.~\Figref{fig:edge_dis} in Section \ref{sec:edge_dis} showcases how the learned edge masking differs for various downstream target links. The frequent occurrence of larger standard deviations of edge maskings of the same edge for different target links implies substantial disagreement on the optimal DAs. This verifies the necessity of our subgraph-based approach for isolating DA effects per target link.
    \item \textbf{Complete before Reduce}: Recognizing that traditional IB-based methods often overlook data instance information recovery prior to compression, we introduced Complete stage. By recovering missing data, the complete stage can attempt to fulfill the critical assumption (2) in Theorem~\ref{thm:reduce} so that the final DA is predictive and concise. The performance comparison in Table~\ref{table:main} necessitates the need for the Complete stage.
\end{itemize}
Furthermore, we introduced several improvements to achieve better performance and stability (See the ablation study in Appendix~\ref{app:ab}):

\begin{itemize}
    \item \textbf{Attention Mechanism}: When pruning the noisy edges, GIB/GSAT assumes that knowing the edge representation itself is sufficient to determine whether it is a critical substructure. However, in our link prediction task, one edge may be critical for one target link but not for another. We propose using target link representation to discern the criticality of an edge for the said link with attention mechanism, which contrasts with GSAT's approach. (~\Eqref{eq:att})
    \item \textbf{Edge Label}: To differentiate original edges from those introduced in the complete stage, we have adopted a labeling strategy based on their scores to discern their relative importance. (Appendix~\ref{app:edge_label})
    \item \textbf{Unbiased KL Loss}: In GSAT implementation, the KL loss term is not an unbiased empirical estimator. In a mini-batch $B$ during training, the KL loss term in GSAT is minimized through $\mathbb{E}_{G}[\text{KL}] \approx \frac{\sum_{G \in B} \text{KL}}{\sum_{G \in B} |E_{G}|}$. That is, their minimization treats each edge in the batch of graphs as a data instance. However, the unbiasd estimator should be $\frac{\sum_{G \in B} \text{KL}}{|B|},$, which average across the subgraphs. The KL loss estimator in GSAT is not that troublesome because the number of edges in their benchmark graphs is similar. However, the number of edges in the subgraphs of the target link varies a lot. The performance comparison in the ablation study necessitates the unbiased KL loss estimator.
\end{itemize}

\paragraph{Distinct Utility}
\begin{itemize}
    \item \textbf{Transferability of Graph Structures}: Our method goes beyond previous IB-based graph ML approaches by verifying that our refined graph structures can serve as inputs to other link prediction models like CN, AA or RA. (See Section 4.2)
    \item \textbf{Robustness to Adversarial Attack}: Our method exhibits robustness to adversarial perturbations targeted at link prediction tasks, further validating the capability of our method to capture critical substructures. (See Section 4.2)
\end{itemize}

In essence, while GSAT inspired our work, our contributions are substantial, addressing nuances of the link prediction task and refining the DA approach for better performance. The ablation studies (Appendix~\ref{app:ab}) suggest that applying IB upon link prediction task is a non-trivial work. It requires a dedicated design to suit the unique challenge of link prediction. These distinctions highlight \our's value and uniqueness in the graph learning domain.

\section{Implementation details}
\label{app:implement}
\subsection{Marginal distribution}
\label{app:marginal}
We discuss the marginal distribution term $r(G_{(i,j)}^{\pm})$ in~\Eqref{eq:reg}.
Since $\tilde{G}_{(i,j)}$ is sampled based on $G_{(i,j)}^{+}$ through $\tilde{\omega}_{(u,v)} \sim \text{Bern}(\gamma)$,
we can write $r(G_{(i,j)}^{\pm}) = \sum_{G_{(i,j)}^{+}}\mathbb{P}(\tilde{\omega}|G_{(i,j)}^{+})\mathbb{P}(G_{(i,j)}^{+})$.
Because $\tilde{\omega}$ is independent from the the inflated graph $G_{(i,j)}^{+}$ given its size $n^\pm$,
$r(G_{(i,j)}^{\pm}) = \sum_{n^\pm}\mathbb{P}(\tilde{\omega}|n^\pm)\mathbb{P}(n^\pm) = \mathbb{P}(n^\pm)\prod_{u,v} \mathbb{P}(\tilde{\omega}_{(u,v)})$.
Thus, the KL-divergence term in~\Eqref{eq:loss} can be written as:
\begin{align}
    &\text{KL}(p_{\phi}(G_{(i,j)}^{\pm} | G_{(i,j)}^{+})||r(G_{(i,j)}^{\pm}))  = \nonumber \\
    &\sum_{(u,v)\in E_{(i,j)}^{+}} p_{(u,v)} \log \frac{p_{(u,v)}}{\gamma} + \left(1-p_{(u,v)}\right) \log \frac{1-p_{(u,v)}}{1-\gamma} \nonumber\\
    &+ \text{Constant},
\end{align}
where $p_{(u,v)}=\text{sigmoid}(a_{(u,v)})$ and the constant term accounts for the terms of node pairs $(u,v)\notin E_{(i,j)}^{+}$ without any trainable parameters. 

In practice, we further allow the constraint hyperparameter $\gamma$ to be different for the original edges in the inflated graph $G_{(i,j)}^{+}$
and those added in the Complete stage. Namely, we have $\gamma_{\text{ori}}$ and $\gamma_{\text{ext}}$ such that:
\begin{align}
    &\text{KL}(p_{\phi}(G_{(i,j)}^{\pm} | G_{(i,j)}^{+})||r(G_{(i,j)}^{\pm})) = \nonumber \\ 
    &\sum_{(u,v)\in E_{(i,j)}^{+} \cap E} p_{(u,v)} \log \frac{p_{(u,v)}}{\gamma_{\text{ori}}} + \left(1-p_{(u,v)}\right) \log \frac{1-p_{(u,v)}}{1-\gamma_{\text{ori}}} \nonumber \\ 
    &+ \sum_{(u,v)\in E_{(i,j)}^{+} \cap E_{ext}} p_{(u,v)} \log \frac{p_{(u,v)}}{\gamma_{\text{ext}}} + \left(1-p_{(u,v)}\right) \log \frac{1-p_{(u,v)}}{1-\gamma_{\text{ext}}} \nonumber\\
    &+ \text{Constant}.
\end{align}

\subsection{Awareness of edges scores from the Complete stage}
\label{app:edge_label}
During the Reduce stage, when obtaining the edge representation, $\mathbf{h}{(u,v)}$, 
we enrich this representation by appending an additional encoding, $\tilde{\mathbf{h}}{(u,v)}$. 
This supplementary encoding serves to inform the model about the edge's origin — 
whether it's a component of the original graph or an edge added in the Complete stage. 
However, this encoding strategy does not provide insights into the relative importance of the newly added edges.

To address this, we incorporate a ranking mechanism, assigning a rank label to each added edge based on its relative importance. 
Specifically, we segregate the added edges into ten equal-sized buckets, determined by their respective scores.
Each edge is then assigned a label corresponding to its bucket number.
This method of score-aware edge representation enables the model to make more informed use of the added edges, and to discern the most informative edges from the others.

\subsection{Augmentation during inference}
During the training phase,~\our reduces the inflated graph by implementing edge sampling. 
However, for the testing phase, we do not employ sampling to obtain the augmented graph.

Given that the entire model can be conceptualized as a probabilistic model, 
the inference stage requires us to compute the expectation of the random variables. 
As such, for each edge $(u,v)$, we use its expected value, $p_{(u,v)}$, as the edge weight during the inference process. 

\subsection{Nodewise sampling}
The implementation of~\our applies an attention mechanism on each edge $(u,v)$ to get $a_{(u,v)}$, which can consume
a huge amount of GPU memory when operating on large-scale graphs. 
As a compromised modification, we follow~\cite{miao_interpretable_2022} to sample the node in the inflated graph instead of the edges.
More specifically, after we get the node representation, we apply the attention to the node and the subgraph representation
to get the importance score $a_{u}$ for each node. Then, each node is still sampled through a Bernoulli distribution $\omega_{u}\sim \text{Bern}(\text{sigmoid}(a_{u}))$. In the end, the edge mask is obtained by $\omega_{(u,v)}=\omega_{u}*\omega_{v}$.

\kevin{weighted graph version of CN AA RA}


\subsection{Hyperparameter details}
In the Complete stage of our experiments, we employ \emph{GCN} and \emph{SAGE} as link predictors to 
inject potential missing edges into the four non-attributed graphs. 
This choice is driven by the smaller sizes of these graphs. 
The parameter $k$, representing the number of top-scored edges, is searched within the range $[1000,2000]$.

For the remaining four attributed graphs, we limit our search to heuristic link predictors, 
namely \emph{CN}, \emph{AA}, and \emph{RA}. The $k$ parameter for these graphs is explored within the range $[16000, 32000]$.

In the Reduce stage, we conduct a hyperparameter search for both $\gamma$ and $\beta$. 
Specifically, we search for the parameters $\gamma_{\text{ori}}$ and $\gamma_{\text{ext}}$ within the set $[0.8, 0.5, 0.2]$. 
The parameter $\beta$ is searched within the range $[1,0.1,0.01]$.

\subsection{Software and hardware details}
Our implementation leverages the PyTorch Geometric library~\cite{fey_fast_2019} and the SEAL~\cite{zhang_link_2018} framework. 
All experiments were conducted on a Linux system equipped with an NVIDIA P100 GPU with 16GB of memory.

\subsection{Time complexity}
The time complexity of our method is primarily similar to that of SEAL. 
However, there are two additional computational requirements: (1) an extra node representation encoding is needed for edge pruning; and (2) a probability score must be assigned to each edge. 
Consequently, the overall computation complexity of our method is $O(t(d^{l+1}F'' + d^{l+1}F''^2))$, where $t$ represents the number of target links, $d$ is the maximum node degree, $l$ corresponds to the number of hops of the subgraph, and $F''$ indicates the dimension of the representation.

\section{Supplementary experiments}
\subsection{Baseline methods}
\label{app:baseline}

\paragraph{\emph{CN}~\cite{liben-nowell_link_2003}.} Common Neighbor (CN) is a widely-used link predictor that posits a node pair with more common neighbors is more likely to connect. The score is computed as $CN(i,j) = |\mathcal{N}_i \cap \mathcal{N}_j|$.

\paragraph{\emph{AA}~\cite{adamic_friends_2003}.} Adamic-Adar (AA) extends the CN approach, emphasizing that common neighbors with fewer connections are more important than those with many connections. The score is calculated as $AA(i,j) = \sum_{z \in \mathcal{N}_i \cap \mathcal{N}_j} \frac{1}{\log |\mathcal{N}_z|}$.

\paragraph{\emph{RA}~\cite{zhou_predicting_2009}.} Resource Allocation (RA) modifies the weight decay of AA on common neighbors. It computes the score as $RA(i,j) = \sum_{z \in \mathcal{N}_i \cap \mathcal{N}_j} \frac{1}{|\mathcal{N}_z|}$.

\paragraph{\emph{GCN}~\cite{kipf_semi-supervised_2017}.} Graph Convolutional Networks (GCN) propose a graph convolution operation using the first-degree neighbors. Owing to its computational efficiency and high performance, GCN is a popular architecture for GNNs.

\paragraph{\emph{SAGE}~\cite{hamilton_inductive_2018}.} GraphSAGE (SAGE) proposes a scalable approach to applying GNNs on large graphs. The encoder part of SAGE uses two distinct weights for the center node representation and its surrounding neighbors.

\paragraph{\emph{GIN}\cite{xu_how_2018}.} Graph Isomorphism Network (GIN) is a 1-WL expressive GNN widely used for graph classification problems. In our experiment, we use the zero-one labeling trick\cite{zhang_labeling_2021} in GIN to distinguish between the target node pair and the remaining nodes in the target link's local neighborhood.

\paragraph{\emph{SEAL}\cite{zhang_link_2018}.} SEAL is a state-of-the-art link prediction model. It uses GNNs and a node labeling trick\cite{zhang_labeling_2021} to enhance expressiveness for link prediction. We found that SEAL's expected performance on the \textbf{Collab} dataset is approximately 5\% higher than initially reported.

\paragraph{\emph{CFLP}~\cite{zhao_learning_2021}.} CFLP introduces counterfactual links into the graph to enable causal inference for link prediction. Despite its efficiency for smaller graphs, CFLP can cause out-of-memory issues for larger graphs due to its preprocessing step to find counterfactual node pairs.

\paragraph{\emph{Edge Proposal}~\cite{singh_edge_2021}.} Edge Proposal augments the graph by adding potential missing or future links to complement the graph for enhanced link prediction performance.

\paragraph{\emph{Node Drop}~\cite{papp_dropgnn_2021}.} Node Drop, also known as DropGNN, randomly drops nodes in the graph. This exposes the model to multiple views of the graph, enhancing its expressiveness.

\paragraph{\emph{Edge Drop}~\cite{rong_dropedge_2020}.} Edge Drop, also known as DropEdge in their work, introduces a stochastic approach to edge removal as a regularization method to solve the oversmoothing issue in GNNs.

\subsection{Benchmark datasets}
\label{app:benchmark_data}
The following graph datasets were utilized in our experiments:

\textbf{Non-attributed graph datasets:}
\begin{enumerate}
\item \textbf{USAir}\cite{batagelj_pajek_2006}: This dataset contains a representation of US airlines, encapsulating the connectivity between different airports.
\item \textbf{Yeast}\cite{von_mering_comparative_2002}: This dataset includes a protein-protein interaction network within yeast cells, providing insights into the complex interplay of biological components.
\item \textbf{C.ele}\cite{watts_collective_1998}: This dataset represents the neural network of the nematode Caenorhabditis elegans, one of the most studied organisms in neuroscience.
\item \textbf{Router}\cite{spring_measuring_2002}: This dataset encompasses Internet connectivity at the router-level, providing a snapshot of the web's underlying infrastructure.
\end{enumerate}

\textbf{Attributed graph datasets:}
\begin{enumerate}
\item \textbf{CS}\cite{shchur_pitfalls_2019}: This dataset provides a snapshot of the collaboration network in the computer science domain, highlighting co-authorship relationships.
\item \textbf{Physics}\cite{shchur_pitfalls_2019}: This dataset depicts a collaboration network within the field of physics, offering insights into academic partnerships.
\item \textbf{Computers}\cite{shchur_pitfalls_2019}: This dataset presents a segment of the co-purchase network on Amazon, reflecting purchasing behavior related to computer products.
\item \textbf{Collab}\cite{wu_graph_2020}: This dataset presents a large-scale collaboration network, showcasing a wide array of interdisciplinary partnerships.

\end{enumerate}

Comprehensive statistics for these datasets are detailed in \autoref{table:dataset}.
Note that when we perform the train test split, we ensure that the split is the same for all different methods on each dataset.

\begin{table*}[h]
\caption{Statistics of benchmark datasets.}
\label{table:dataset}
            \begin{center}
            \resizebox{1\textwidth}{!}{
            \begin{tabular}{lcccccc}
                \toprule
                \textbf{Dataset} & \textbf{\#Nodes} & \textbf{\#Edges} & \textbf{Avg. node deg.} & \textbf{Max. node deg.} & \textbf{Density} & \textbf{Attr. Dimension}\\
                \midrule
                \textbf{C.ele} & 297 & 4296 & 14.46 & 134 & 9.7734\% & - \\
                \textbf{USAir} & 332 & 4252 & 12.81 & 139 & 7.7385\% & - \\
                \textbf{Yeast} & 2375 & 23386 & 9.85 & 118 & 0.8295\% & - \\
                \textbf{Router} & 5022 & 12516 & 2.49 & 106 & 0.0993\% & - \\
                \textbf{CS} & 18333 & 163788 & 8.93 & 136 & 0.0975\% & 6805 \\
                \textbf{Physics} & 34493 & 495924 & 14.38 & 382 & 0.0834\% & 8415 \\
                \textbf{Computers} & 13752 & 491722 & 35.76 & 2992 & 0.5201\% & 767 \\
                \textbf{Collab} & 235868 & 2358104 & 10.00 & 671 & 0.0085\% & 128 \\
            \bottomrule
            \end{tabular}
            }
            \end{center}
            \end{table*}

\begin{table*}[h]
    \centering
\caption{Results 
of adversarial robustness for different models on the rest of datasets. 
The attack rates of 10\%, 30\%, and 50\% represent the respective ratios of edges subjected to adversarial flips by CLGA~\cite{zhang_unsupervised_2022}.}\label{table:robust2}
\resizebox{1\linewidth}{!}{
\begin{tabular}{c|cccc|cccc|cccc|cccc|cccc|cccc|cccc}
    \toprule[1.5pt]
    \multirow{2}{*}{\textbf{Methods}} & \multicolumn{4}{c|}{Yeast} & \multicolumn{4}{c|}{Router} & \multicolumn{4}{c|}{CS} & \multicolumn{4}{c|}{Physics} & \multicolumn{4}{c|}{Computers}\\
    &No Adv&10\%&30\%&50\%&No Adv&10\%&30\%&50\%&No Adv&10\%&30\%&50\%&No Adv&10\%&30\%&50\%&No Adv&10\%&30\%&50\%\\\midrule
    \emph{GCN} & 80.33 & 76.69 & 68.30 & 57.96 & 35.16 & 28.55 & 20.75 & 15.86 & 60.69 & 64.24 & 57.49 & 45.28 & 69.16 & 68.85 & 60.16 & 46.76 & 32.70 & 30.16 & 24.33 & 20.07\\ 
    \emph{SAGE} & 78.34 & 74.33 & 67.01 & 56.23 & 35.76 & 35.13 & 29.01 & 25.11 & 31.44 & 53.92 & 42.72 & 29.16 & 22.87 & 61.69 & 50.20 & 36.83 & 14.53 & 10.42 & 4.16 & 4.34\\ 
    \emph{ELPH} & 78.92 & 76.74 & 69.05 & 65.58 & 59.50 & 57.07 & 52.83 & 47.65 & 67.84 & 65.92 & 60.28 & 50.09 & 69.60 & 64.67 & 58.83 & 47.51 & 33.64 & 34.30 & 29.35 & 23.23\\ 
    \emph{NCNC} & 73.11 & 71.34 & 66.34 & 59.60 & 57.13 & 55.04 & 52.01 & 47.47 & 65.73 & 63.13 & 53.93 & 36.66 & 72.87 & 69.27 & 58.74 & 45.31 & 37.17 & 35.74 & 33.59 & 31.53\\ 
    \emph{SEAL} & 82.50 & 78.24 & 69.24 & 62.84 & 60.35 & 51.84 & 48.75 & 44.02 & 65.23 & 60.31 & 58.97 & 42.38 & 71.83 & 64.28 & 59.84 & 44.17 & 35.80 & 33.84 & 31.84 & 28.84\\ 
    \emph{CORE} & \textbf{84.67} & \textbf{81.94} & \textbf{74.70} & \textbf{68.96} & \textbf{65.64} & \textbf{59.20} & \textbf{56.58} & \textbf{51.02} & \textbf{69.67} & \textbf{66.87} & \textbf{61.18} & \textbf{50.49} & \textbf{74.73} & \textbf{70.78} & \textbf{62.58} & \textbf{50.37} & \textbf{37.88} & \textbf{36.28} & \textbf{34.66} & \textbf{31.85}\\
  \bottomrule[1.5pt]
\end{tabular}
}
\end{table*}

\subsection{More ablation study}
\label{app:ab}
\begin{table*}[h]
\caption{Ablation study evaluated by Hits@50. 
The \textbf{best-performing} components are highlighted in bold, while the \underline{second-best} performance is underlined.}
\label{table:ablation2}
\begin{center}
  \resizebox{1\textwidth}{!}{
  \begin{NiceTabular}{lcccccccc}
    \toprule
    \textbf{Models} & \textbf{C.ele} & \textbf{USAir} & \textbf{Router} & \textbf{Yeast} & \textbf{CS} & \textbf{Physics} & \textbf{Computers} & \textbf{Collab}\\
    
    \midrule
    \small{No Attention (~\Eqref{eq:att})} & 74.41{\tiny$\pm$1.75} & \underline{92.38{\tiny$\pm$1.07}} & 63.60{\tiny$\pm$2.92} & 82.81{\tiny$\pm$0.94} & 65.69{\tiny$\pm$2.38} & \underline{73.10{\tiny$\pm$0.78}} & 36.85{\tiny$\pm$1.17} & \underline{70.09{\tiny$\pm$0.94}} \\
    \small{No Edge Label (Appendix~\ref{app:edge_label})} & 74.27{\tiny$\pm$4.73} & 91.49{\tiny$\pm$0.81} & 63.07{\tiny$\pm$3.90} & \underline{83.79{\tiny$\pm$2.01}} & \underline{65.84{\tiny$\pm$1.40}} & 71.71{\tiny$\pm$2.72} & 36.79{\tiny$\pm$2.88} & 69.64{\tiny$\pm$1.32} \\
    \small{Biased KL Loss} & \underline{75.14{\tiny$\pm$2.18}} & 90.66{\tiny$\pm$3.69} & \underline{64.80{\tiny$\pm$2.25}} & 81.91{\tiny$\pm$1.70} & 64.22{\tiny$\pm$3.38} & 69.32{\tiny$\pm$5.05} & \underline{37.31{\tiny$\pm$2.47}} & 69.61{\tiny$\pm$1.13} \\
    \small{CORE} & \textbf{76.34{\tiny$\pm$1.65}} & \textbf{92.69{\tiny$\pm$0.75}} & \textbf{65.47{\tiny$\pm$2.44}} & \textbf{84.22{\tiny$\pm$1.58}} & \textbf{68.15{\tiny$\pm$0.78}} & \textbf{74.73{\tiny$\pm$2.12}} & \textbf{37.88{\tiny$\pm$1.10}} & \textbf{70.64{\tiny$\pm$0.51}} \\
  \bottomrule
\end{NiceTabular}
}
\end{center}
\end{table*}

We further conduct ablation studies on three unique components we specifically design for link prediction tasks, to enhance the performance of our proposed methods. The results are presented in Table~\ref{table:ablation2}. As the results suggest, all three components can boost the performance of the proposed DA method by varying degrees. For instance,~\our with \textit{No Attention} hampers the performance from $0.6$ to $2.5$ in Hits@50. ~\our with \textit{No Edge Label} also drops the effectiveness of DAs by from $0.5$ to $3$. The \textit{Biased KL Loss} mostly hurts the performance of~\our on \textbf{USAir} and \textbf{Physics}, which we assume that the degree distribution of these two datasets is more skewed compared to others.

\subsection{Parameter sensitivity}
\label{app:param_sen}
\begin{figure*}
\begin{center}
\centerline{\includegraphics[width=\textwidth]{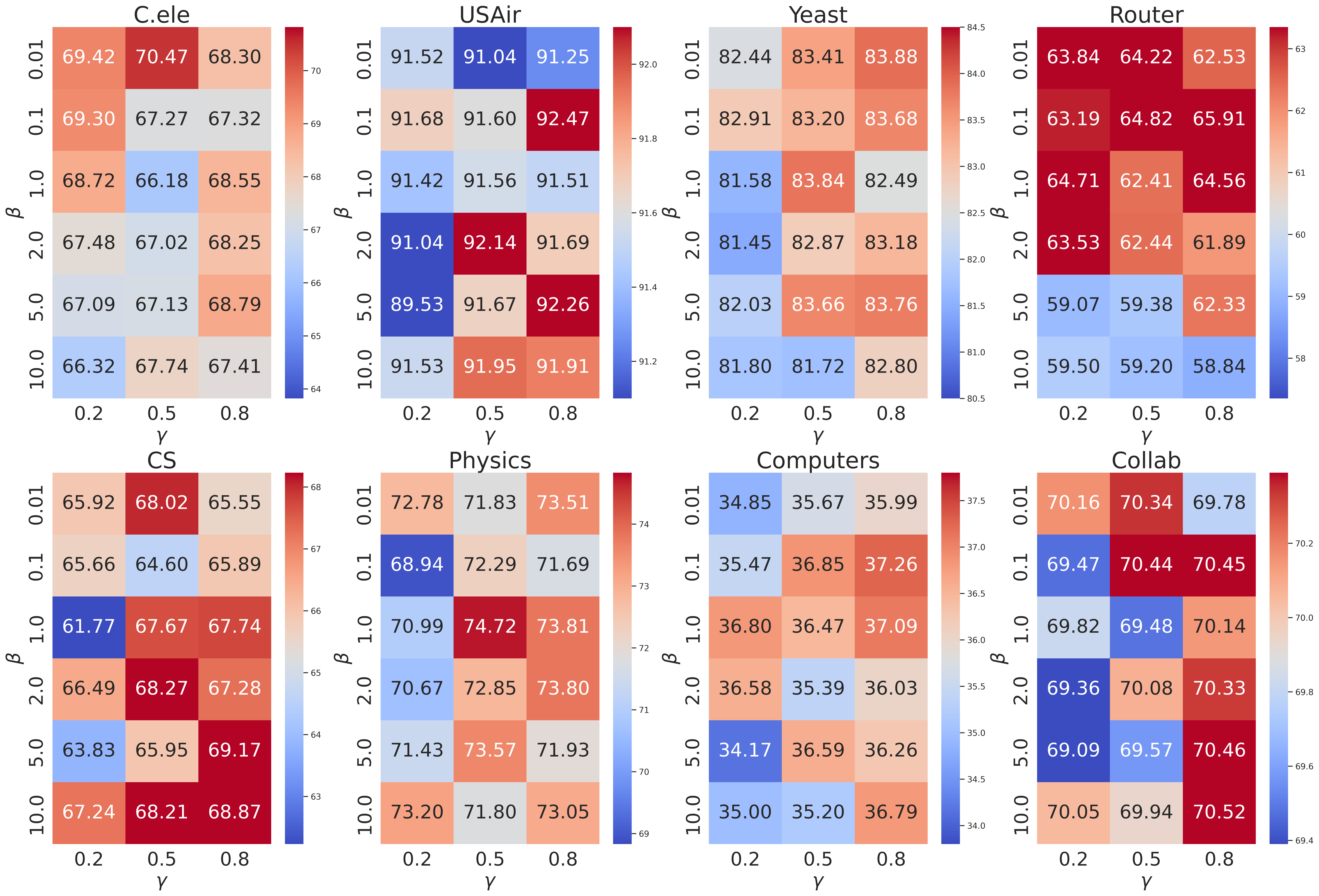}}
\caption{\our can improve LP performance in various hyperparameter settings measured by Hits@50.
Warmer colors indicate improved performance over the baseline, whereas cooler colors signify the contrary.}\label{fig:heatmap}
\end{center}
\end{figure*}

We also conducted an experiment to examine the sensitivity of the hyperparameters in \our. 
We focused on the Reduce stage only, as this is the core component of our method. 
As per our hyperparameter search procedure, we evaluated the model performance for $\beta$ across $[10,5,2,1,0.1,0.01]$ and $\gamma$ across $[0.8,0.5,0.2]$. 
The results are depicted in~\Figref{fig:heatmap}.
Our method consistently enhanced the model performance across various hyperparameter settings. However, we also notice that the performance of \our will drop if we increase the graph compression term $\beta$ or decrease the edge-preserving term $\gamma$. This is because sparsifying the graph too much will result in the loss of critical information. Thus, balanced $\beta$ and $\gamma$ are crucial for a robust \our performance.

Besides, the parameter $\gamma$ acts as a regulatory mechanism that influences each edge's probability, nudging it towards the behavior of a random graph. A lower $\gamma$ tends to suppress non-essential edges in predictions. Based on our observations, values between $[0.5, 0.8]$ tend to be optimal. That being said, in our studies, the balancing factor $\beta$ has exhibited a more pronounced effect on model performance than $\gamma$.

\subsection{\our with GCN and SAGE as backbones}
\label{app:gcn_sage}

\begin{table}[h]
    \centering
\caption{\our with GCN and SAGE as backbone models. The \textbf{best-performing} methods are highlighted in bold, while the \underline{second-best} performance is underlined.}\label{table:gcn_sage}
\resizebox{1\linewidth}{!}{
\begin{tabular}{l cccc}
                \toprule
                \textbf{Methods} & \textbf{C.ele} & \textbf{USAir} & \textbf{Router} & \textbf{Yeast}\\
                
                \midrule
                \multicolumn{5}{l}{\emph{GCN} as the backbone model} \\
                \emph{GCN} & 62.21{\tiny$\pm$6.13} & 83.20{\tiny$\pm$3.88} & 43.37{\tiny$\pm$9.75} & 81.30{\tiny$\pm$1.96} \\
                \emph{Complete Only} & 65.20{\tiny$\pm$3.76} & 85.84{\tiny$\pm$1.62} & \underline{53.18{\tiny$\pm$6.92}} & 81.54{\tiny$\pm$1.10} \\
                \emph{Reduce Only} & \underline{65.52{\tiny$\pm$2.95}}& \underline{86.42{\tiny$\pm$4.03}} & 47.26{\tiny$\pm$8.37} & \underline{82.82{\tiny$\pm$0.96}} \\ 
                \emph{CORE} & \textbf{68.79{\tiny$\pm$2.48}} & \textbf{87.96{\tiny$\pm$1.03}} & \textbf{55.35{\tiny$\pm$4.53}} & \textbf{82.82{\tiny$\pm$0.96}} \\
                \midrule
                \multicolumn{5}{l}{\emph{SAGE} as the backbone model} \\
                \emph{SAGE} & 70.91{\tiny$\pm$1.05} & 80.38{\tiny$\pm$7.18} & 56.71{\tiny$\pm$2.59} & 84.70{\tiny$\pm$2.01} \\
                \emph{Complete Only} & 71.59{\tiny$\pm$2.15} & 83.60{\tiny$\pm$2.98} & 58.60{\tiny$\pm$2.79} & 84.91{\tiny$\pm$1.33} \\
                \emph{Reduce Only} & \underline{72.12{\tiny$\pm$1.84}} & \underline{87.32{\tiny$\pm$3.83}} & \underline{59.54{\tiny$\pm$2.69}} & \underline{85.47{\tiny$\pm$0.96}} \\ 
                \emph{CORE} & \textbf{73.36{\tiny$\pm$2.08}} & \textbf{89.76{\tiny$\pm$2.25}} & \textbf{61.75{\tiny$\pm$1.07}} & \textbf{85.61{\tiny$\pm$0.98}} \\
              \bottomrule
            \end{tabular}
}
\end{table}

\begin{table}[h]
    \centering
\caption{Number of node pairs with at least one common neighbor as a positive instance in the testing sets.}\label{table:one_common}
\resizebox{1\linewidth}{!}{
\begin{tabular}{lccc}
                \toprule
                \textbf{Dataset} & \# Node pairs with CN  & \# Node pairs & Ratio\\
                \midrule
                \textbf{C.ele} & 344 & 429 & 80.17\%\\
                \textbf{USAir} & 387 & 425 & 91.05\%\\
                \textbf{Yeast} & 1700 & 2338 & 72.71\%\\
                \textbf{Router} & 114 & 1251 & 9.11\%\\
                \textbf{CS} & 11306&16378&69.03\% \\
                \textbf{Physics} & 42061 & 49592 &84.81\% \\
                \textbf{Computers} & 45676&49172&92.89\%\\
            \bottomrule
            \end{tabular}
}
\end{table}
We further investigate~\our with GCN and SAGE as backbones. The results are presented in Table~\ref{table:gcn_sage}. It shows that~\our can consistently improve model performance with various GNNs as the backbones.

\subsection{Complete stage only considering node pairs with common neighbors}
\label{app:only_common}
When we score the potential node pairs to be added into the graph at Complete stage, we only consider those with at least one common neighbor. While this seems to limit the capability of recovering missing edges, it is actually an effective approach with balanced computational efficiency. We empirically investigate the number of node pairs in the testing set that have at least one common neighbor, presented in Table~\ref{table:one_common}.

With the exception of the Router dataset, our benchmarks consistently indicate that positive testing edges are inclined to have at least one common neighbor. Therefore, our scoring process at the Complete stage encompasses the majority of the testing edges. This observation underscores that our edge addition strategy aligns well with the community-like nature observed in real-world datasets.

\section{Variational bounds for the GIB objective in~\Eqref{eq:pred} and~\Eqref{eq:reg}}
\label{app:bound}
The objective from~\Eqref{eq:GIB} is:
\begin{equation}
    -I(G_{(i,j)}^{\pm},Y) + \beta I(G_{(i,j)}^{\pm},G_{(i,j)}^{+}). \nonumber
\end{equation}
Since those two terms are computationally intractable, we introduce two variational bounds.

For $I(G_{(i,j)}^{\pm},Y)$, by the definition of mutual information:
\begin{align*}  
    I(G_{(i,j)}^{\pm},Y) &= \mathbb{E} [ \log \frac{p(Y|G_{(i,j)}^{\pm})}{p(Y)} ] \\
    &=\mathbb{E} [ \log \frac{q_{\theta}(Y|G_{(i,j)}^{\pm})}{ p(Y)} ] \\
    &+ \mathbb{E}[\text{KL}(p(Y|G_{(i,j)}^{\pm})  || q_{\theta}(Y|G_{(i,j)}^{\pm}))] \\
    & \geq \mathbb{E} [ \log \frac{q_{\theta}(Y|G_{(i,j)}^{\pm})}{ p(Y)} ]  \\
    & = \mathbb{E}[ \log q_{\theta}(Y|G_{(i,j)}^{\pm}) ] + H(Y),
\end{align*}
where the KL-divergence term is nonnegative.

For the second term $I(G_{(i,j)}^{\pm},G_{(i,j)}^{+})$, by definition:
\begin{align*}
    I(G_{(i,j)}^{\pm},G_{(i,j)}^{+}) &= \mathbb{E} [\log \frac{p(G_{(i,j)}^{\pm}|G_{(i,j)}^{+})}{p(G_{(i,j)}^{\pm})}] \\
    &= \mathbb{E} [\log \frac{p_{\phi}(G_{(i,j)}^{\pm}|G_{(i,j)}^{+})}{r(G_{(i,j)}^{\pm})}] \\
    &- \mathbb{E}[\text{KL} ( p(G_{(i,j)}^{\pm}) || r(G_{(i,j)}^{\pm}) )]  \\
    &\leq \mathbb{E}[\text{KL}(p_{\phi}(G_{(i,j)}^{\pm} | G_{(i,j)}^{+})||r(G_{(i,j)}^{\pm})].
\end{align*}

\section{Proof for Theorem~\ref{thm:reduce}}
\label{app:proof}
We restate Theorem~\ref{thm:reduce}:
Assume that: \text{\normalfont (1)} The existence $Y$ of a link $(i,j)$ is solely determined 
by its local neighborhood
$G_{(i,j)}^{*}$ in a way such that $p(Y) = f(G_{(i,j)}^{*})$,
where $f$ is a deterministic invertible function;
\text{\normalfont (2)} The inflated graph contains sufficient structures for prediction
$G_{(i,j)}^{*} \in \mathbb{G}_{\text{sub}}(G_{(i,j)}^{+})$.
Then $G_{(i,j)}^{\pm} = G_{(i,j)}^{*}$ minimizes the objective in~\Eqref{eq:GIB}.

\allowdisplaybreaks
\begin{proof}
We can follow a similar derivation as in~\cite{miao_interpretable_2022}. Consider the following steps:

\begin{align*}
& - I(G_{(i,j)}^{\pm};Y) + \beta I(G_{(i,j)}^{\pm};G_{(i,j)}^{+}) \\ \tag{Start with the original objective~\Eqref{eq:GIB}} \\
= & - I(G_{(i,j)}^{+},G_{(i,j)}^{\pm};Y) + I(G_{(i,j)}^{+};Y|G_{(i,j)}^{\pm}) + \beta I(G_{(i,j)}^{\pm};G_{(i,j)}^{+}) \\ \tag{Expand the first term via chain rule of mutual information} \\
= & - I(G_{(i,j)}^{+},G_{(i,j)}^{\pm};Y) + (1-\beta) I(G_{(i,j)}^{+};Y|G_{(i,j)}^{\pm}) \\
&+ \beta I(G_{(i,j)}^{+};G_{(i,j)}^{\pm},Y) \\ \tag{Split the third term proportionally} \\
= & - I(G_{(i,j)}^{+};Y) + (1-\beta) I(G_{(i,j)}^{+};Y|G_{(i,j)}^{\pm}) + \beta I(G_{(i,j)}^{+};G_{(i,j)}^{\pm},Y) \\ \tag{Because $G_{(i,j)}^{\pm}$ is a subgraph of $G_{(i,j)}^{+}$} \\ 
= & (\beta-1) I(G_{(i,j)}^{+};Y) - (\beta-1) I(G_{(i,j)}^{+};Y|G_{(i,j)}^{\pm}) \\
&+ \beta I(G_{(i,j)}^{+};G_{(i,j)}^{\pm}|Y), \\ \tag{Split the last term and rearrange terms}
\end{align*}

Since $I(G_{(i,j)}^{+};Y)$ does not involve trainable parameters, we focus on the last two terms. 
If $\beta \in [0,1]$, the $G_{(i,j)}^{\pm}$ that minimizes~\Eqref{eq:GIB} also minimizes $-(\beta-1) I(G_{(i,j)}^{+};Y|G_{(i,j)}^{\pm}) + \beta I(G_{(i,j)}^{+};G_{(i,j)}^{\pm}|Y)$. 
Given that mutual information is nonnegative,
the lower bound of $(1-\beta) I(G_{(i,j)}^{+};Y|G_{(i,j)}^{\pm}) + \beta I(G_{(i,j)}^{+};G_{(i,j)}^{\pm}|Y)$ is 0.

Next, we show that $G_{(i,j)}^{\pm} = G_{(i,j)}^{*}$ can make~\Eqref{eq:GIB} reach its lower bound.
Since $p(Y) = f(G_{(i,j)}^{*})$, $I(G_{(i,j)}^{+};Y|G_{(i,j)}^{*})=0$.
That is, there is no mutual information between $G_{(i,j)}^{+}$ and $Y$ when knowing $G_{(i,j)}^{+};Y$.
Similarly, because $f$ is invertible, there is no more mutual information between $G_{(i,j)}^{+}$ and $G_{(i,j)}^{\pm}$ when knowing $Y$,
yielding $I(G_{(i,j)}^{+};G_{(i,j)}^{\pm}|Y)=0$.
Therefore, $G_{(i,j)}^{\pm} = G_{(i,j)}^{*}$ minimizes~\Eqref{eq:GIB}.
\end{proof}

\section{Limitations}
In this section, we address the limitations of our proposed method. 
Firstly, while our model,~\our, delivers superior performance through the application of distinct augmentations for each link prediction, 
this practice significantly increases the computational burden. 
This is due to the requirement of independently calculating the augmentation for each link. 
Attempts to implement a universal augmentation across all links simultaneously resulted in a significant performance drop. 
Thus, future work may explore efficient methods to balance computational overhead with performance gains.

Secondly, our backbone model, SEAL, utilizes node labeling to determine proximity to the target link for nodes within the neighborhood. 
Due to computational constraints, this labeling process is performed on the CPU during preprocessing. 
Despite the capacity of the Reduce stage to alter the local structure of the target link, 
the node labels remain unchanged post-graph pruning, potentially leading to information leakage about each node's position in the unaltered graph. 
Future research could investigate methods to fully conceal this information, enabling link prediction to be purely dependent on the pruned graph.
\clearpage

\end{document}